\documentclass[twoside]{article}

\usepackage{xcolor}
\definecolor{grey}{rgb}{0.5,0.5,0.5}

\usepackage[round]{natbib}

\usepackage[preprint]{aistats2026}

\makeatletter
\def\ps@plain{%
  \let\@mkboth\@gobbletwo
  \let\@oddhead\@empty
  \let\@evenhead\@empty
  \def\@oddfoot{\hfil\thepage\hfil}%
  \def\@evenfoot{\hfil\thepage\hfil}%
}
\makeatother
\usepackage{graphicx}
\usepackage{amsmath}
\usepackage{amssymb}
\usepackage{amsfonts}
\usepackage{booktabs}
\usepackage{placeins}
\usepackage{float}

\definecolor{steelblue}{RGB}{70,130,180}
\usepackage[
    colorlinks=true,
    linkcolor=black,
    citecolor=black,
    urlcolor=steelblue
]{hyperref}

\begin{document}

\runningtitle{DR-SNE}
\runningauthor{Kazanskii}

\twocolumn[

\aistatstitle{DR-SNE: Density-Regularized Stochastic Neighbor Embedding}

\aistatsauthor{Maksim Kazanskii$^{*}$}

\aistatsaddress{
Independent Researcher \\
$^{*}$\texttt{mkazanskii@gmail.com}
}
]

\begin{abstract}
Dimensionality-reduction methods such as t-SNE preserve local neighborhood structure but can substantially distort the local distribution of data. We introduce Density-Regularized Stochastic Neighbor
Embedding (DR-SNE), which augments stochastic neighbor embedding with a
regularizer that directly aligns normalized inverse-neighborhood-scale
profiles between the original and embedding spaces. The resulting
objective preserves relative local concentration while remaining
invariant to global rescaling of the embedding. Across real and synthetic datasets, DR-SNE generally improves preservation of the targeted concentration profile while maintaining controlled levels of neighborhood fidelity. 
Experiments reveal a consistent empirical trade-off between concentration preservation and neighborhood fidelity as the regularization strength varies. These results position DR-SNE as a simple extension of SNE
for settings in which relative variation in local concentration is an
important property of the representation.
\end{abstract}

\section{Introduction}

Many dimensionality-reduction methods formulate embedding construction as a
geometric problem: given high-dimensional data, the goal is to preserve
informative relationships in a lower-dimensional representation.
Methods such as t-SNE~\citep{maaten2008tsne},
UMAP~\citep{mcinnes2018umap},
TriMap~\citep{amid2019trimap}, and
PaCMAP~\citep{wang2021pacmap} preserve geometric structure through
different objectives. t-SNE matches pairwise neighborhood probabilities
between the original and embedding spaces~\citep{maaten2008tsne};
UMAP constructs a weighted nearest-neighbor graph and optimizes a
low-dimensional fuzzy-set representation of this graph~\citep{mcinnes2018umap};
TriMap uses triplet constraints to preserve relative relationships among
observations~\citep{amid2019trimap}; and PaCMAP optimizes pairwise relations
at multiple scales using near, mid-near, and further point
pairs~\citep{wang2021pacmap}. Despite these differences, these objectives
do not explicitly require preservation of relative local neighborhood scale.
Preserving neighborhood relationships does not necessarily
preserve how local neighborhood scale varies across observations:
relatively compact regions may become dispersed, while diffuse regions may
become artificially compact
\citep{wattenberg2016tsne,narayan2021density}.

Density-aware extensions such as DensMAP and DenSNE~\citep{narayan2021density} explicitly address aspects of local density preservation. However, neighborhood structure, relative local concentration, and dimensionally calibrated probability density represent distinct properties of the data. Neighborhood structure
describes which observations remain locally related, whereas relative
local concentration describes how an observation's characteristic
neighborhood scale compares with those of other observations.
Conventional $k$-NN density estimators additionally depend on
neighborhood volume and dimensionality, while an inverse characteristic
neighborhood length measures local concentration without directly
estimating probability density.

We therefore focus on preserving relative local concentration. For each
observation, we characterize neighborhood by an inverse characteristic
length scale and normalize these statistics across the dataset. The
resulting profile identifies observations lying in relatively compact or
dispersed regions. Preserving this profile can prevent embeddings from
artificially homogenizing variation in local scale and may retain
information useful to downstream procedures based on neighborhood
geometry.

We introduce Density-Regularized SNE (DR-SNE), an extension of stochastic neighbor embedding that explicitly regularizes this relative local-concentration profile. Here, “density” refers to relative local sample concentration rather than to a dimensionally calibrated probability-density estimate. DR-SNE aligns normalized
inverse-neighborhood-scale statistics in log-space between the original
and embedding spaces while retaining the neighborhood-preserving SNE
objective. As in prior density-aware approaches, normalization removes sensitivity
to global scale. DR-SNE differs in directly aligning the resulting
normalized relative local-concentration profiles pointwise rather than
encouraging correlation between local-scale profiles.

\section{Related Work}

Classical dimensionality-reduction methods preserve different geometric
properties: PCA preserves dominant variance directions~\citep{jolliffe2002pca},
Isomap approximates manifold geometry through geodesic
distances~\citep{tenenbaum2000isomap}, and LLE preserves local linear
reconstruction relationships~\citep{roweis2000lle}.
\textit{t}-distributed stochastic neighbor embedding
(t-SNE)~\citep{maaten2008tsne} is widely used due to its strong local
neighborhood preservation and ability to reveal cluster structure, with
several extensions improving scalability and
optimization~\citep{linderman2019fast}.
However, t-SNE is known to distort global geometry and can distort
variations in data density, often producing embeddings with misleading
cluster sizes and spatial relationships~\citep{wattenberg2016tsne}.
Uniform Manifold Approximation and Projection
(UMAP)~\citep{mcinnes2018umap} provides an alternative graph-based
formulation based on a fuzzy representation of neighborhood structure.
Other dimensionality-reduction methods such as
TriMap~\citep{amid2019trimap} and PaCMAP~\citep{wang2021pacmap} use
alternative objective formulations designed to better balance local and
global structure.

Density-aware variants such as DensMAP and
DenSNE~\citep{narayan2021density} incorporate information about local
crowding by encouraging consistency between neighborhood-scale quantities
across spaces. DR-SNE is closely related to these approaches but differs
in the criterion used to preserve this information. DenSNE and DensMAP
encourage correlation between log local-radius profiles, whereas DR-SNE
directly aligns a dataset-normalized relative local-concentration profile
pointwise in log-space. This distinction is consequential:
correlation-based alignment encourages linear association between
local-scale profiles but is insensitive to positive affine transformations
of either profile, whereas the DR-SNE objective remains sensitive to
systematic amplification or compression of relative local-scale variation.
At the same time, normalization makes DR-SNE invariant to a global
multiplicative change in neighborhood scale.

Thus, DR-SNE should not be interpreted as replacing the local-scale
quantities of prior density-aware methods with a probability-density
estimator. Rather, it introduces a different preservation criterion:
direct pointwise alignment of a normalized relative-concentration profile
instead of correlation-based agreement between local-radius profiles.
We evaluate the relationship between this targeted property and
conventional density preservation separately using a conventional
$k$-NN volume-density estimator.
\section{Methods}
\subsection{Density-Regularized Stochastic Neighbor Embedding}

Given $X=\{x_i\}_{i=1}^n\subset\mathbb{R}^D$, we seek an embedding
$Z=\{z_i\}_{i=1}^n\subset\mathbb{R}^d$ ($d\ll D$) that preserves
local neighborhoods while controlling distortions of relative local
concentration.

Following t-SNE~\citep{maaten2008tsne}, we define
\begin{equation}
p_{ij}
=
\frac{p_{j|i}+p_{i|j}}{2n},
\qquad
p_{j|i}\propto
\exp(-\beta_i\|x_i-x_j\|^2),
\end{equation}
where $\beta_i$ is set by the specified perplexity, and
\begin{equation}
q_{ij}
=
\frac{(1+\|z_i-z_j\|^2)^{-1}}
{\sum_{k\neq l}(1+\|z_k-z_l\|^2)^{-1}}.
\end{equation}

While t-SNE preserves pairwise affinities, it does not explicitly align
a prescribed pointwise statistic of neighborhood scale between the two
spaces. DR-SNE addresses this limitation by augmenting the neighborhood
objective with an explicit relative local-concentration regularizer.
The relationship between these two objectives, including the KL
decomposition and invariance properties, is analyzed in
Appendix~\ref{app:theory}.

For neighborhood preservation, we use a sparse $k$-NN KL objective,
\begin{equation}
\mathcal{L}_{\mathrm{KL}}
=
\sum_{i=1}^n\sum_{j\in\mathcal{N}_k(i)}
p_{ij}\log\frac{p_{ij}}{q_{ij}}.
\end{equation}
\paragraph{Relative local concentration.}
For each observation, we characterize local concentration by the inverse
mean distance to its $k$ nearest neighbors:
\begin{equation}
c_i^{\mathrm{high}}
=
\left(
\frac{1}{k}\sum_{j\in\mathcal{N}_k(i)}
\|x_i-x_j\|+\epsilon
\right)^{-1},
\label{eq:concentration_high}
\end{equation}
\begin{equation}
c_i^{\mathrm{low}}
=
\left(
\frac{1}{k}\sum_{j\in\mathcal{N}_k(i)}
\|z_i-z_j\|+\epsilon
\right)^{-1}.
\label{eq:concentration_low}
\end{equation}
Here, $\epsilon>0$ ensures numerical stability. Larger values indicate
more compact local neighborhoods. Unlike conventional $k$-NN
volume-density estimators, which scale approximately as $r_{i,k}^{-D}$
and $r_{i,k}^{-d}$ in the original and embedding spaces, respectively,
$c_i$ is an inverse characteristic length scale and is not intended as
a dimensionally calibrated probability-density estimate.

DR-SNE instead preserves the \emph{relative concentration profile}.
To remove global scale dependence, we normalize each profile to unit mean:
\begin{equation}
\tilde{c}_i^{\mathrm{high}}
=
\frac{c_i^{\mathrm{high}}}
{\frac{1}{n}\sum_{\ell}c_\ell^{\mathrm{high}}},
\qquad
\tilde{c}_i^{\mathrm{low}}
=
\frac{c_i^{\mathrm{low}}}
{\frac{1}{n}\sum_{\ell}c_\ell^{\mathrm{low}}},
\end{equation}
We align the normalized profiles in log-space:
\begin{equation}
\mathcal{L}_{\mathrm{conc}}
=
\frac{1}{n}\sum_{i=1}^{n}
\left(
\log\tilde{c}_i^{\mathrm{high}}
-
\log\tilde{c}_i^{\mathrm{low}}
\right)^2,
\label{eq:concentration_loss}
\end{equation}
so that the penalty reflects multiplicative differences in relative
local concentration rather than absolute neighborhood scale.

The final objective is
\begin{equation}
\mathcal{L}
=
\begin{cases}
\mathcal{L}_{\mathrm{KL}}^{\mathrm{early}},
& t<T_{\mathrm{warmup}},\\[2pt]
\mathcal{L}_{\mathrm{KL}}
+\lambda\mathcal{L}_{\mathrm{conc}},
& \text{otherwise}.
\end{cases}
\end{equation}
Early exaggeration is applied during warm-up. We optimize with Adam,
gradient clipping, and Gaussian initialization; additional implementation
details are provided in Appendix~\ref{sec:implementation_details}.

Preservation under a conventional $k$-NN volume-density estimator is evaluated separately.

\subsection{Comparison Methods}

We compare DR-SNE with t-SNE~\citep{maaten2008tsne},
UMAP~\citep{mcinnes2018umap}, PaCMAP~\citep{wang2021pacmap},
and the density-aware methods DensMAP and
DenSNE~\citep{narayan2021density}. We additionally include a
high-dimensional baseline (HIGH-D), in which evaluations are performed
directly in the original feature space, providing a reference for the
information retained by the embeddings.

\section{Results}

\subsection{Visualization and Experiments}

We evaluate embeddings using
\textit{trustworthiness}~\citep{venna2001neighborhood} to quantify local
neighborhood preservation, \textit{concentration correlation} (introduced later) to measure
preservation of the targeted relative local-concentration profile, and
\textit{$k$-NN density rank correlation} ($\rho_S$) as an objective-independent
measure of conventional density preservation. 

We quantify preservation of the targeted relative local-concentration
profile using \emph{concentration correlation} (CC),
\begin{equation}
\mathrm{CC}
=
\mathrm{corr}
\left(
\log \tilde{c}_i^{(\mathrm{high})},
\log \tilde{c}_i^{(\mathrm{low})}
\right).
\end{equation}
CC directly measures preservation of the relative local-concentration
statistic targeted by DR-SNE and is therefore an objective-aligned
diagnostic rather than an independent measure of probability-density
preservation.

To evaluate whether preservation of local concentration also transfers to a conventional notion of density, we additionally report an objective-independent $k$-NN volume-density metric. We estimate local density using the standard $k$-NN volume estimator,
\begin{equation}
\hat{p}_{k}(x_i)
\propto
r_{i,k}^{-d},
\end{equation}
where $r_{i,k}$ is the distance to the $k$th nearest neighbor and $d$
is the dimensionality of the corresponding space. We report the
Spearman rank correlation between the resulting density estimates in
the original and embedded spaces (\emph{kNN Density} $\rho_S$).

Unless stated otherwise, all evaluation metrics use fixed,
method-independent evaluation settings. This separates three distinct
questions: whether an embedding preserves local neighborhoods, whether
it preserves the relative local-concentration statistic explicitly
targeted by DR-SNE, and whether this preservation generalizes to
conventional density estimators.
\begin{table*}[!t]
\centering
\scriptsize
\setlength{\tabcolsep}{4pt}
\renewcommand{\arraystretch}{1.05}

\begin{tabular}{lcccccc}
\toprule
 & \textbf{t-SNE} 
 & \textbf{UMAP} 
 & \textbf{PaCMAP} 
 & \textbf{DensMAP} 
 & \textbf{DenSNE} 
 & \textbf{DR-SNE} \\
\midrule

% ============================================================
% DIGITS
% ============================================================
\multicolumn{7}{c}{\textbf{Digits}} \\
\cmidrule(lr){1-7}

TW $\uparrow$
& \textbf{0.980 $\pm$ 0.002}
& 0.967 $\pm$ 0.001
& 0.953 $\pm$ 0.002
& 0.954 $\pm$ 0.003
& 0.975 $\pm$ 0.004
& 0.959 $\pm$ 0.009 \\

Concentration Corr. $\uparrow$
& 0.628 $\pm$ 0.020
& 0.554 $\pm$ 0.002
& 0.559 $\pm$ 0.008
& 0.840 $\pm$ 0.002
& 0.779 $\pm$ 0.011
& \textbf{0.922 $\pm$ 0.029} \\

kNN Density $\rho_S$ $\uparrow$
& 0.795 $\pm$ 0.015
& 0.546 $\pm$ 0.017
& 0.517 $\pm$ 0.109
& 0.865 $\pm$ 0.006
& \textbf{0.905 $\pm$ 0.004}
& 0.460 $\pm$ 0.157 \\

% ============================================================
% FASHION-MNIST
% ============================================================
\addlinespace[5pt]
\midrule
\multicolumn{7}{c}{\textbf{Fashion-MNIST}} \\
\midrule

TW $\uparrow$
& \textbf{0.973 $\pm$ 0.000}
& 0.952 $\pm$ 0.000
& 0.956 $\pm$ 0.001
& 0.955 $\pm$ 0.001
& 0.961 $\pm$ 0.001
& 0.959 $\pm$ 0.004 \\

Concentration Corr. $\uparrow$
& 0.481 $\pm$ 0.003
& 0.358 $\pm$ 0.003
& 0.383 $\pm$ 0.013
& 0.750 $\pm$ 0.003
& 0.671 $\pm$ 0.068
& \textbf{0.820 $\pm$ 0.061} \\

kNN Density $\rho_S$ $\uparrow$
& 0.485 $\pm$ 0.017
& 0.076 $\pm$ 0.005
& 0.065 $\pm$ 0.040
& 0.776 $\pm$ 0.002
& \textbf{0.907 $\pm$ 0.004}
& 0.505 $\pm$ 0.105 \\

% ============================================================
% MNIST
% ============================================================
\addlinespace[5pt]
\midrule
\multicolumn{7}{c}{\textbf{MNIST}} \\
\midrule

TW $\uparrow$
& \textbf{0.941 $\pm$ 0.007}
& 0.921 $\pm$ 0.000
& 0.921 $\pm$ 0.000
& 0.890 $\pm$ 0.000
& 0.930 $\pm$ 0.001
& 0.893 $\pm$ 0.011 \\

Concentration Corr. $\uparrow$
& 0.118 $\pm$ 0.004
& 0.039 $\pm$ 0.010
& 0.120 $\pm$ 0.003
& 0.987 $\pm$ 0.000
& 0.908 $\pm$ 0.010
& \textbf{0.991 $\pm$ 0.004} \\

kNN Density $\rho_S$ $\uparrow$
& -0.003 $\pm$ 0.011
& -0.002 $\pm$ 0.008
& \textbf{0.007 $\pm$ 0.009}
& -0.011 $\pm$ 0.007
& -0.009 $\pm$ 0.002
& -0.006 $\pm$ 0.010 \\

% ============================================================
% PBMC
% ============================================================
\addlinespace[5pt]
\midrule
\multicolumn{7}{c}{\textbf{PBMC}} \\
\midrule

TW $\uparrow$
& 0.879 $\pm$ 0.001
& 0.859 $\pm$ 0.001
& 0.859 $\pm$ 0.001
& 0.852 $\pm$ 0.000
& \textbf{0.880 $\pm$ 0.004}
& 0.859 $\pm$ 0.002 \\

Concentration Corr. $\uparrow$
& 0.218 $\pm$ 0.000
& 0.032 $\pm$ 0.008
& 0.273 $\pm$ 0.004
& 0.824 $\pm$ 0.005
& 0.820 $\pm$ 0.005
& \textbf{0.974 $\pm$ 0.002} \\

kNN Density $\rho_S$ $\uparrow$
& 0.361 $\pm$ 0.029
& 0.056 $\pm$ 0.041
& 0.103 $\pm$ 0.012
& 0.831 $\pm$ 0.002
& \textbf{0.919 $\pm$ 0.012}
& 0.749 $\pm$ 0.021 \\

% ============================================================
% SHUTTLE
% ============================================================

\addlinespace[5pt]
\midrule
\multicolumn{7}{c}{\textbf{Shuttle}} \\
\midrule

TW $\uparrow$
& \textbf{0.998 $\pm$ 0.000}
& 0.996 $\pm$ 0.000
& 0.995 $\pm$ 0.000
& 0.991 $\pm$ 0.002
& 0.993 $\pm$ 0.002
& 0.992 $\pm$ 0.001 \\

Concentration Corr. $\uparrow$
& 0.274 $\pm$ 0.016
& 0.216 $\pm$ 0.006
& 0.154 $\pm$ 0.010
& 0.663 $\pm$ 0.034
& 0.347 $\pm$ 0.022
& \textbf{0.894 $\pm$ 0.039} \\

kNN Density $\rho_S$ $\uparrow$
& 0.078 $\pm$ 0.019
& -0.018 $\pm$ 0.012
& -0.098 $\pm$ 0.006
& \textbf{0.770 $\pm$ 0.011}
& 0.733 $\pm$ 0.044
& 0.594 $\pm$ 0.060 \\

% ============================================================
% CONTROLLED CLUSTERS
% ============================================================
\addlinespace[5pt]
\midrule
\multicolumn{7}{c}{\textbf{Controlled Clusters (synthetic)}} \\
\midrule

TW $\uparrow$
& \textbf{0.997 $\pm$ 0.002}
& 0.969 $\pm$ 0.009
& 0.961 $\pm$ 0.010
& 0.953 $\pm$ 0.001
& 0.991 $\pm$ 0.002
& 0.965 $\pm$ 0.013 \\

Concentration Corr. $\uparrow$
& 0.165 $\pm$ 0.096
& 0.100 $\pm$ 0.029
& 0.083 $\pm$ 0.037
& 0.975 $\pm$ 0.001
& 0.399 $\pm$ 0.080
& \textbf{0.978 $\pm$ 0.011} \\

kNN Density $\rho_S$ $\uparrow$
& 0.201 $\pm$ 0.196
& 0.071 $\pm$ 0.059
& 0.047 $\pm$ 0.078
& 0.893 $\pm$ 0.003
& \textbf{0.932 $\pm$ 0.004}
& 0.768 $\pm$ 0.020 \\

% ============================================================
% DENSITY-GRADIENT HELIX
% ============================================================
\addlinespace[5pt]
\midrule
\multicolumn{7}{c}{\textbf{Concentration-Gradient Helix (synthetic)}} \\
\midrule

TW $\uparrow$
& 0.996 $\pm$ 0.001
& \textbf{0.999 $\pm$ 0.000}
& 0.997 $\pm$ 0.001
& 0.986 $\pm$ 0.008
& 0.996 $\pm$ 0.001
& 0.967 $\pm$ 0.005 \\

Concentration Corr. $\uparrow$
& 0.484 $\pm$ 0.045
& 0.371 $\pm$ 0.021
& 0.466 $\pm$ 0.017
& 0.970 $\pm$ 0.004
& 0.525 $\pm$ 0.026
& \textbf{0.995 $\pm$ 0.001} \\

kNN Density $\rho_S$ $\uparrow$
& 0.533 $\pm$ 0.063
& 0.649 $\pm$ 0.048
& 0.327 $\pm$ 0.034
& \textbf{0.907 $\pm$ 0.003}
& 0.894 $\pm$ 0.009
& 0.243 $\pm$ 0.091 \\

\bottomrule
\end{tabular}

\vspace{4pt}

\caption{
Comparison across datasets.
TW denotes trustworthiness.
Concentration Corr.\ measures preservation of the normalized
inverse-neighborhood-scale statistic targeted by DR-SNE.
To independently assess preservation of conventional density structure,
we additionally report Spearman rank correlation ($\rho_S$) between
high- and low-dimensional density estimates obtained using a
$k$-nearest-neighbor volume estimator ($k=30$).
The $k$-NN density metric is not used by the DR-SNE optimization
objective or for hyperparameter selection.
Best values for each metric and dataset are highlighted in bold.
Results are reported as mean $\pm$ standard deviation across
independent random seeds.
}
\label{tab:main_comparison}
\end{table*}
To enable a controlled comparison, we adopt a constrained evaluation protocol. For each dataset, hyperparameters are selected to maximize concentration correlation subject to a common minimum trustworthiness constraint. This ensures that all selected configurations satisfy a prescribed level of local neighborhood preservation while allowing trustworthiness to vary above that threshold.  In effect, we analyze the trade-off between neighborhood fidelity and relative local-concentration preservation by asking which method better preserves the targeted concentration profile among configurations satisfying the prescribed trustworthiness constraint. This provides a constrained comparison of the trade-off between neighborhood preservation and local concentration/density preservation.

\begin{figure*}[p]
\centering
\includegraphics[width=\textwidth,height=0.9\textheight,keepaspectratio]{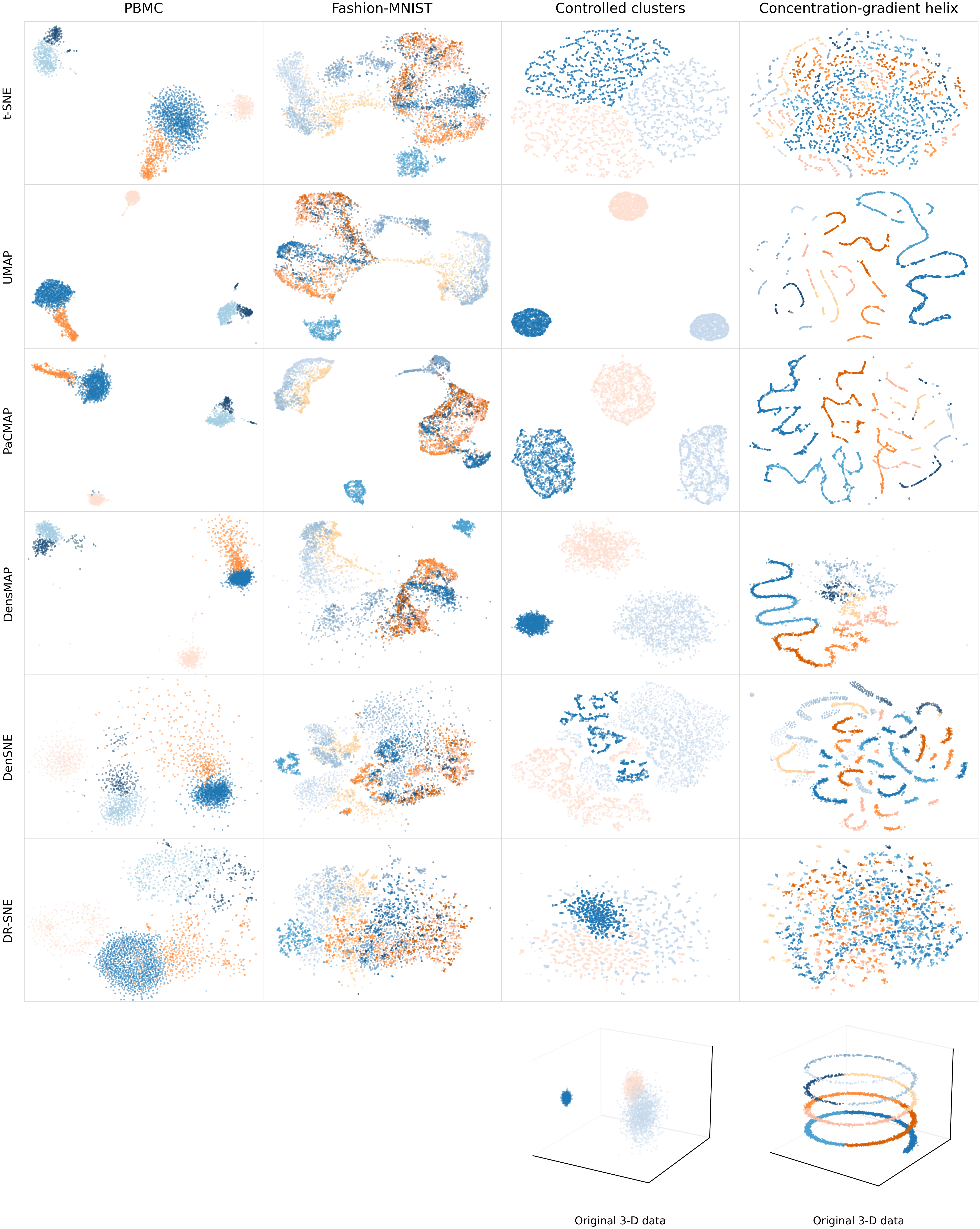}
\caption{
Comparison of dimensionality reduction methods on real-world and
synthetic datasets. Rows show t-SNE, UMAP, PaCMAP, DensMAP, DenSNE,
and DR-SNE; columns show PBMC, Fashion-MNIST, Controlled Clusters,
and Concentration-Gradient Helix. Original three-dimensional data are
shown below the corresponding synthetic embedding columns.
}
\label{fig:main_real_world}
\end{figure*}
We evaluate the qualitative behavior of the methods on two real-world datasets and two controlled synthetic benchmarks. The real-world evaluation covers both biological and image data. \textbf{PBMC}~\citep{zheng2017massively} is a single-cell RNA-seq dataset representing heterogeneous immune-cell populations; we normalize and log-transform the counts, select the 2,000 most highly variable genes, scale the features, and apply PCA before embedding. \textbf{Fashion-MNIST}~\citep{xiao2017fashion} consists of $28\times28$ grayscale images from ten clothing categories; we randomly sample 5,000 observations and standardize the features. To complement these real-world datasets, we use two synthetic benchmarks in which variation in local concentration is controlled by construction. \textbf{Controlled Clusters} consists of three Gaussian populations with equal sample counts but different spatial scales, providing a direct test of whether embeddings preserve relative differences between compact and diffuse regions. \textbf{Concentration-Gradient Helix} consists of points sampled non-uniformly along a three-dimensional helix, producing a continuous variation in local concentration along the manifold. Full construction procedures and parameters for both synthetic datasets are provided in Appendix~\ref{sec:synthetic_datasets}. Additional qualitative comparisons on
\textbf{Digits}~\citep{pedregosa2011scikit},
\textbf{MNIST}~\citep{lecun1998mnist}, and
\textbf{Shuttle}~\citep{uci_shuttle}
are provided in Appendix~\ref{sec:visualization}.

\begin{table*}[!t]
\centering
\scriptsize
\setlength{\tabcolsep}{5pt}
\renewcommand{\arraystretch}{1.15}

\caption{
Ablation over concentration regularization strength $\lambda$ on the PBMC dataset.
Results are mean $\pm$ standard deviation over 3 runs.
}
\label{tab:lambda_ablation_pbmc_full}

\begin{tabular}{lcccccc}
\toprule
$\lambda$
& Trustworthiness
& Continuity
& Concentration correlation
& Silhouette
& Stress
& Time (s) \\
\midrule
0.0
& $0.910\pm0.002$
& $0.890\pm0.010$
& $0.470\pm0.037$
& $0.352\pm0.064$
& $3.002\pm0.020$
& $21.26$ \\

0.001
& $\mathbf{0.925\pm0.002}$
& $\mathbf{0.922\pm0.009}$
& $0.767\pm0.006$
& $\mathbf{0.398\pm0.056}$
& $2.132\pm0.137$
& $20.79$ \\

0.005
& $0.913\pm0.005$
& $0.902\pm0.023$
& $0.927\pm0.004$
& $0.299\pm0.094$
& $2.336\pm0.048$
& $20.27$ \\

0.01
& $0.900\pm0.007$
& $0.882\pm0.032$
& $0.961\pm0.001$
& $0.227\pm0.096$
& $2.604\pm0.118$
& $20.27$ \\

0.05
& $0.835\pm0.005$
& $0.820\pm0.005$
& $0.993\pm0.000$
& $0.055\pm0.005$
& $2.460\pm0.144$
& $20.52$ \\

0.1
& $0.805\pm0.009$
& $0.806\pm0.008$
& $\mathbf{0.997\pm0.000}$
& $0.033\pm0.009$
& $\mathbf{2.229\pm0.137}$
& $21.27$ \\
\bottomrule
\end{tabular}

\end{table*}
Qualitatively, the methods exhibit distinct behaviors in how they balance neighborhood geometry and variation in local concentration (Figure~\ref{fig:main_real_world}). On PBMC, t-SNE, UMAP, and PaCMAP produce relatively compact and well-separated structures, whereas DensMAP and DenSNE retain more variation in local scale; DR-SNE emphasizes this variation most strongly, producing clearly differentiated compact and dispersed regions but less sharply separated clusters. A similar pattern appears on Fashion-MNIST: t-SNE, UMAP, and PaCMAP preserve visually distinct class structure, while DensMAP and DenSNE provide an intermediate behavior between geometric separation and variation in local scale. DR-SNE produces a more diffuse representation in which differences in local concentration remain pronounced, illustrating the cost in cluster separation that can accompany stronger concentration regularization. The synthetic datasets make these differences particularly transparent. On the synthetic benchmarks, where the underlying concentration structure is known by construction, the differences between methods are especially clear. For Controlled Clusters, t-SNE, UMAP, and PaCMAP preserve the separation of the three populations but substantially alter their relative spatial scales. DensMAP preserves the compact-to-diffuse ordering particularly well, while DenSNE shows a weaker correspondence to the original differences in dispersion. DR-SNE also retains a pronounced distinction between the three concentration regimes, consistent with its high concentration correlation. The Concentration-Gradient Helix presents a different challenge because concentration varies continuously along a single manifold. Here, t-SNE, UMAP, and PaCMAP largely preserve local manifold structure but break the helix into multiple separated segments. DensMAP retains both the concentration variation and a comparatively coherent organization of the manifold, whereas DenSNE produces stronger fragmentation. DR-SNE preserves the relative concentration profile particularly strongly, but the two-dimensional embedding no longer retains the global helical organization. Together, these examples show that concentration preservation and manifold geometry need not coincide: DensMAP can achieve a strong balance between the two, whereas DR-SNE more directly favors preservation of the targeted relative concentration profile.

Under the constrained evaluation protocol, DR-SNE attains the highest concentration correlation across all evaluated datasets. However, concentration correlation directly measures the statistic targeted by the DR-SNE objective and should therefore be interpreted primarily as an objective-aligned self-test: these results confirm that the proposed optimization successfully preserves the property it is designed to control, rather than establishing objective-independent superiority. More importantly, this strong concentration preservation is achieved while maintaining competitive trustworthiness across datasets. Performance on the independent $k$-NN density metric is more mixed, with DensMAP and DenSNE often performing better, further demonstrating that the relative local-concentration profile targeted by DR-SNE is distinct from conventional probability-density preservation.

\subsection{Ablation Studies}

We conduct an ablation study over the concentration regularization strength $\lambda$ to assess its impact on embedding quality (Table~\ref{tab:lambda_ablation_pbmc_full}). Experiments are performed on a 3,000-cell subset of the PBMC single-cell RNA-seq dataset using a standard preprocessing pipeline: data are normalized, log-transformed, restricted to the top 2,000 highly variable genes, and scaled, followed by PCA to 30 components. The embedding is constructed from the PCA representation, with a neighborhood size $k = 300$ used for concentration estimation, and perplexity fixed to 30 for computing the similarity matrix.
We observe low variance across three runs, indicating stable optimization behavior.

\begin{figure}[t]
\centering
\includegraphics[width=\columnwidth]{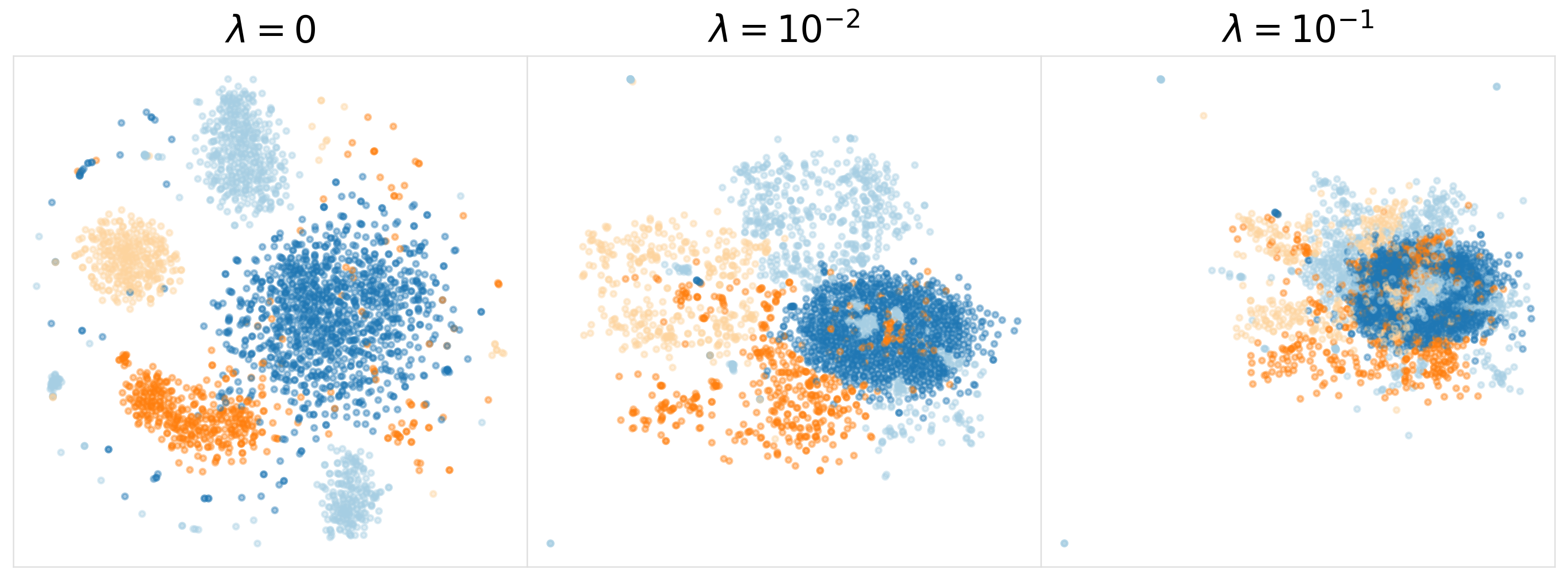}
\caption{
Effect of the concentration regularization parameter $\lambda$ on the embedding.
As $\lambda$ increases (from left to right), concentration preservation improves while the global structure gradually dissipates.
Each panel shows the embedding obtained for a different value of $\lambda$.
}
\label{fig:lambda_progression}
\end{figure}

As $\lambda$ increases, concentration preservation improves substantially,
with concentration correlation rising from $0.47$ at $\lambda=0$ to nearly
$1.0$ at $\lambda=0.1$. The effect on neighborhood preservation is
non-monotonic: the highest mean trustworthiness and continuity occur at the
small non-zero value $\lambda=10^{-3}$, indicating that modest concentration
regularization need not reduce neighborhood fidelity. Under stronger
regularization, however, both metrics decline, with trustworthiness decreasing
from $0.925$ at $\lambda=10^{-3}$ to $0.805$ at $\lambda=0.1$. Silhouette
scores also approach zero, indicating reduced cluster separability. This behavior is consistent with Figure~\ref{fig:lambda_progression}. At
$\lambda=0$, the embedding exhibits clear cluster separation but weak
concentration preservation. At $\lambda=10^{-2}$, relative concentration
variation is more strongly preserved while substantial neighborhood structure
remains. At $\lambda=10^{-1}$, concentration preservation becomes dominant,
accompanied by reduced visual cluster separation.

\begin{figure}[!t]
    \centering
    \includegraphics[width=\columnwidth]{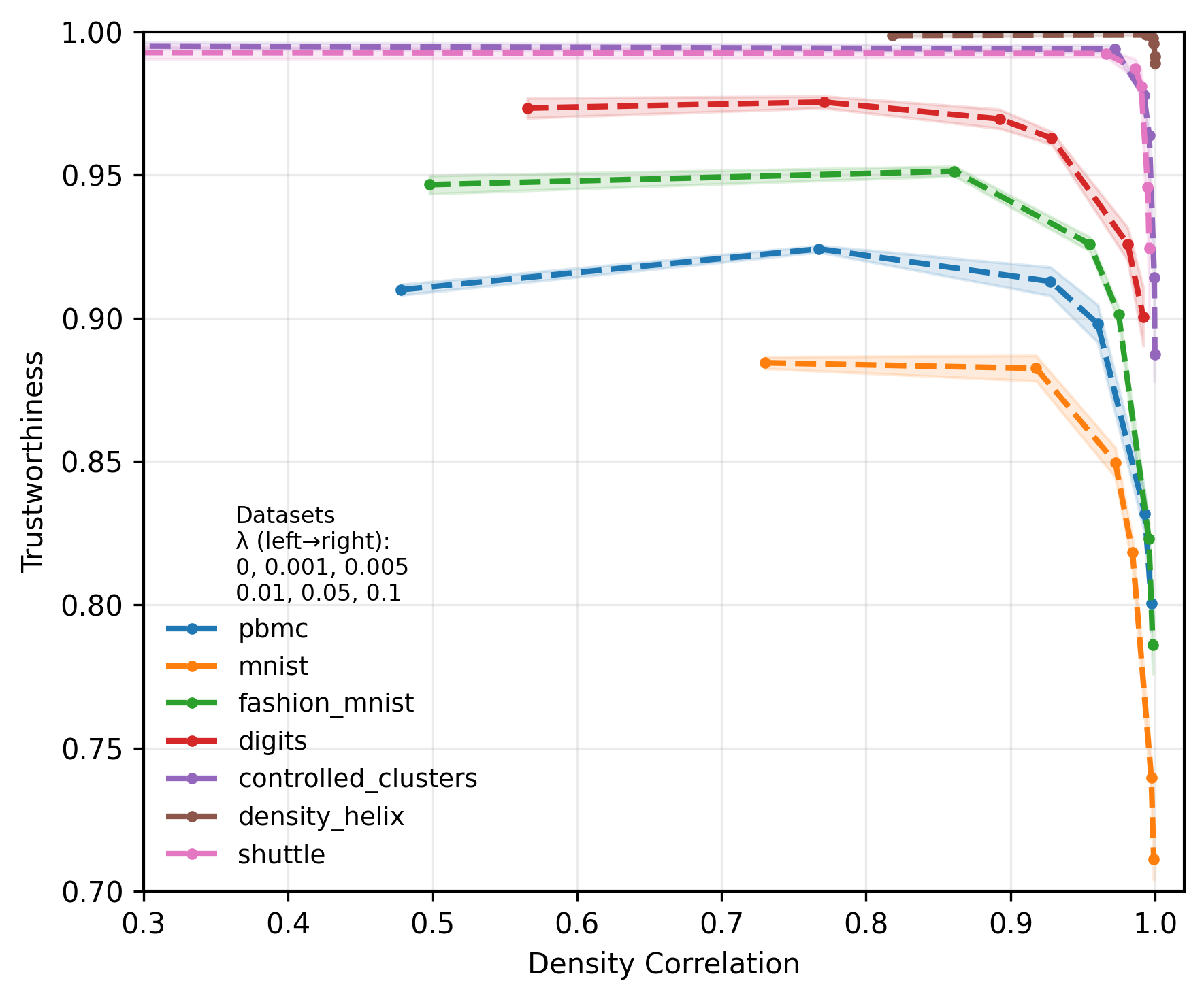}
    \caption{
        Trade-off between trustworthiness and concentration preservation across datasets. Each curve corresponds to a dataset, with points obtained by varying the concentration regularization parameter $\lambda$. Solid lines show the mean across 3 random seeds, and shaded bands indicate $\pm$ one standard deviation. Increasing $\lambda$ generally improves concentration correlation, while trustworthiness often reaches its best value at a small non-zero $\lambda$ before declining at stronger regularization, revealing a broadly consistent empirical trade-off across datasets.
    }
    \label{fig:pareto_lambda}
\end{figure}

To assess whether this trade-off generalizes across datasets, we plot trustworthiness versus concentration correlation (Figure~\ref{fig:pareto_lambda}), where each point corresponds to a different value of $\lambda$. Across datasets, increasing $\lambda$ generally shifts the embeddings toward stronger concentration preservation, with local-structure degradation becoming more evident at larger values. The curves are not strictly monotonic in trustworthiness; small non-zero values of $\lambda$ often yield slight improvements over $\lambda = 0$, suggesting that mild concentration regularization  can enhance local neighborhood preservation. Overall, the trade-off exhibits a broadly similar shape across datasets, with intermediate values of $\lambda$ often occupying favorable regions of the empirical trade-off. 

Additional ablation studies on preprocessing and neighborhood size are
presented in Appendix~\ref{sec:ablation_studies}. In particular, we analyze
the effect of PCA dimensionality prior to embedding, as well as the choice
of the number of nearest neighbors $k$. These experiments assess the
sensitivity of DR-SNE to these design choices and complement the
regularization-strength ablation above.

\subsection{Downstream Evaluation: Anomaly Detection}

To assess whether improved concentration preservation yields practical benefits, we evaluate low-dimensional embeddings on downstream anomaly detection tasks. 
Given a dataset $X$ with binary labels indicating normal and anomalous samples, we first compute embeddings $Z \in \mathbb{R}^2$ for each method across a range of hyperparameter settings.We evaluate on datasets commonly used in anomaly-detection benchmarks
\citep{campos2016evaluation, goldstein2016comparative}
to facilitate comparison with prior work. However, instead of the commonly used Digits dataset, we replace it with the \textbf{Thyroid dataset} from the UCI Machine Learning Repository~\citep{uci_thyroid}, which provides a more realistic biomedical setting. All input features are standardized prior to embedding to ensure a consistent scale across methods. 

\setlength{\textfloatsep}{6pt}

\begin{table*}[!t]
\centering
\scriptsize
\setlength{\tabcolsep}{2.5pt}
\renewcommand{\arraystretch}{1.02}

\caption{
AUPRC (Area Under the Precision--Recall Curve; mean $\pm$ standard deviation) over
5 runs for kNN (k-Nearest Neighbors), LOF (Local Outlier Factor),
IF (Isolation Forest), and Cent (centroid-based detector).
Left: \textbf{Oracle-best} performance, where embedding hyperparameters
are selected independently for each method and detector by maximizing
AUPRC using the ground-truth anomaly labels.
Right: \textbf{Oracle-IF} selection, where a single embedding
configuration for each method is selected by maximizing Isolation Forest
AUPRC and is subsequently evaluated with all four detectors.
Because both selection procedures use ground-truth labels, these results
should be interpreted as label-informed upper-bound analyses rather than
as fully unsupervised model-selection protocols.
Each dataset caption reports the number of samples ($n$) and anomaly
rate (a.r.).
}
\label{tab:anomaly_table_three_main}

% ===================== THYROID =====================
\textbf{Thyroid}, $n = 3771$, $a.r. = 0.077$
\vspace{2pt}

\resizebox{\textwidth}{!}{
\begin{tabular}{lcccc|cccc l}
\toprule
 & \multicolumn{4}{c}{Oracle-best}
 & \multicolumn{4}{c}{Oracle-IF} & \\
\cmidrule(lr){2-5} \cmidrule(lr){6-9}
Method & kNN & LOF & IF & Cent & kNN & LOF & IF & Cent & Params \\
\midrule

\textbf{DR-SNE}
& \textbf{.134 $\pm$ .011}
& .135 $\pm$ .018
& \textbf{.119 $\pm$ .006}
& \textbf{.113 $\pm$ .005}
& .115 $\pm$ .004
& .118 $\pm$ .016
& \textbf{.119 $\pm$ .006}
& \textbf{.113 $\pm$ .006}
& $\lambda$=1.0, k=80 \\

DenSNE
& .133 $\pm$ .008
& .133 $\pm$ .006
& .097 $\pm$ .008
& .091 $\pm$ .005
& .106 $\pm$ .010
& .103 $\pm$ .018
& .097 $\pm$ .008
& .091 $\pm$ .005
& p=5, $\lambda$=0 \\

DensMAP
& .118 $\pm$ .011
& \textbf{.163 $\pm$ .031}
& .079 $\pm$ .004
& .069 $\pm$ .004
& .109 $\pm$ .004
& \textbf{.134 $\pm$ .015}
& .079 $\pm$ .004
& .068 $\pm$ .004
& $\lambda$=10, f=0.3 \\

HIGH-D
& .122 $\pm$ .000
& .124 $\pm$ .000
& .099 $\pm$ .000
& .103 $\pm$ .000
& \textbf{.122 $\pm$ .000}
& .124 $\pm$ .000
& .099 $\pm$ .000
& .103 $\pm$ .000
& -- \\

PaCMAP
& .133 $\pm$ .012
& .107 $\pm$ .007
& \textbf{.119 $\pm$ .004}
& .074 $\pm$ .017
& .084 $\pm$ .002
& .096 $\pm$ .002
& \textbf{.119 $\pm$ .004}
& .070 $\pm$ .000
& k=80, F.Pairs=1 \\

UMAP
& .090 $\pm$ .003
& .106 $\pm$ .011
& .080 $\pm$ .008
& .081 $\pm$ .004
& .083 $\pm$ .003
& .090 $\pm$ .007
& .080 $\pm$ .008
& .081 $\pm$ .004
& k=40, md=0.5 \\

t-SNE
& .108 $\pm$ .003
& .099 $\pm$ .012
& .103 $\pm$ .023
& .083 $\pm$ .005
& .099 $\pm$ .004
& .094 $\pm$ .001
& .103 $\pm$ .023
& .080 $\pm$ .011
& p=50, lr=50 \\

\bottomrule
\end{tabular}
}

\vspace{4pt}

% ===================== CIFAR =====================
\textbf{CIFAR}, $n = 5000$, $a.r. = 0.095$
\vspace{2pt}

\resizebox{\textwidth}{!}{
\begin{tabular}{lcccc|cccc l}
\toprule
 & \multicolumn{4}{c}{Oracle-best}
 & \multicolumn{4}{c}{Oracle-IF} & \\
\cmidrule(lr){2-5} \cmidrule(lr){6-9}
Method & kNN & LOF & IF & Cent & kNN & LOF & IF & Cent & Params \\
\midrule

\textbf{DR-SNE}
& .135 $\pm$ .010
& .107 $\pm$ .005
& .162 $\pm$ .029
& .173 $\pm$ .045
& .135 $\pm$ .010
& .107 $\pm$ .005
& .162 $\pm$ .029
& .173 $\pm$ .045
& $\lambda$=5e-05,k=40 \\

DenSNE
& \textbf{.150 $\pm$ .017}
& .122 $\pm$ .009
& .155 $\pm$ .020
& .171 $\pm$ .092
& \textbf{.150 $\pm$ .017}
& \textbf{.121 $\pm$ .006}
& .155 $\pm$ .020
& .155 $\pm$ .014
& p=5, $\lambda$=0.1 \\

DensMAP
& .138 $\pm$ .006
& .115 $\pm$ .014
& .137 $\pm$ .024
& .096 $\pm$ .031
& .136 $\pm$ .006
& .111 $\pm$ .009
& .137 $\pm$ .024
& .092 $\pm$ .031
& $\lambda$=5, f=0.3 \\

HIGH-D
& .126 $\pm$ .000
& .115 $\pm$ .000
& .121 $\pm$ .000
& .124 $\pm$ .000
& .126 $\pm$ .000
& .115 $\pm$ .000
& .121 $\pm$ .000
& .124 $\pm$ .000
& -- \\

PaCMAP
& .128 $\pm$ .016
& .115 $\pm$ .007
& \textbf{.175 $\pm$ .062}
& \textbf{.222 $\pm$ .171}
& .120 $\pm$ .005
& .113 $\pm$ .006
& \textbf{.175 $\pm$ .062}
& \textbf{.222 $\pm$ .171}
& k=5, F.Pairs=5 \\

UMAP
& .134 $\pm$ .006
& \textbf{.123 $\pm$ .006}
& .141 $\pm$ .027
& .084 $\pm$ .005
& .111 $\pm$ .004
& .095 $\pm$ .003
& .141 $\pm$ .027
& .084 $\pm$ .005
& k=15, md=0 \\

t-SNE
& .146 $\pm$ .013
& .116 $\pm$ .005
& .153 $\pm$ .038
& .194 $\pm$ .207
& .146 $\pm$ .013
& .109 $\pm$ .004
& .153 $\pm$ .038
& .124 $\pm$ .048
& p=100, lr=500 \\

\bottomrule
\end{tabular}
}

\vspace{4pt}

% ===================== SHUTTLE =====================
\textbf{Shuttle}, $n = 5000$, $a.r. = 0.224$
\vspace{2pt}

\resizebox{\textwidth}{!}{
\begin{tabular}{lcccc|cccc l}
\toprule
 & \multicolumn{4}{c}{Oracle-best}
 & \multicolumn{4}{c}{Oracle-IF} & \\
\cmidrule(lr){2-5} \cmidrule(lr){6-9}
Method & kNN & LOF & IF & Cent & kNN & LOF & IF & Cent & Params \\
\midrule

\textbf{DR-SNE}
& \textbf{.521 $\pm$ .051}
& .329 $\pm$ .016
& .603 $\pm$ .047
& .619 $\pm$ .034
& .247 $\pm$ .042
& .261 $\pm$ .005
& .603 $\pm$ .047
& .619 $\pm$ .034
& $\lambda$=5e-04, k=300 \\

DenSNE
& .481 $\pm$ .020
& .292 $\pm$ .019
& .616 $\pm$ .120
& .648 $\pm$ .076
& .379 $\pm$ .004
& .211 $\pm$ .002
& .616 $\pm$ .120
& .634 $\pm$ .083
& p=100, $\lambda$=0.2 \\

DensMAP
& .484 $\pm$ .009
& \textbf{.330 $\pm$ .005}
& .690 $\pm$ .131
& .700 $\pm$ .198
& .182 $\pm$ .006
& .240 $\pm$ .009
& .690 $\pm$ .131
& .666 $\pm$ .193
& $\lambda$=0.01, f=0.5 \\

HIGH-D
& .418 $\pm$ .000
& .299 $\pm$ .000
& .642 $\pm$ .000
& .609 $\pm$ .000
& \textbf{.418 $\pm$ .000}
& \textbf{.299 $\pm$ .000}
& .642 $\pm$ .000
& .609 $\pm$ .000
& -- \\

PaCMAP
& .205 $\pm$ .024
& .278 $\pm$ .008
& \textbf{.841 $\pm$ .014}
& \textbf{.911 $\pm$ .006}
& .141 $\pm$ .002
& .258 $\pm$ .005
& \textbf{.841 $\pm$ .014}
& \textbf{.852 $\pm$ .006}
& k=40, F.Pairs=5 \\

UMAP
& .233 $\pm$ .020
& .251 $\pm$ .011
& .773 $\pm$ .043
& .757 $\pm$ .069
& .174 $\pm$ .008
& .243 $\pm$ .006
& .773 $\pm$ .043
& .588 $\pm$ .115
& k=60, md=0 \\

t-SNE
& .317 $\pm$ .075
& .258 $\pm$ .009
& .698 $\pm$ .054
& .699 $\pm$ .077
& .167 $\pm$ .004
& .241 $\pm$ .004
& .698 $\pm$ .054
& .615 $\pm$ .127
& p=75, lr=500 \\

\bottomrule
\end{tabular}
}

\end{table*}
We then apply a set of anomaly scoring functions directly in the embedding
space. Specifically, we consider: (i) $k$-nearest neighbor (kNN) distance,
defined as the sum of distances to the $k$ nearest neighbors
\citep{ramaswamy2000efficient}; (ii) Local Outlier Factor (LOF), which
measures deviations in local density relative to neighboring points
\citep{breunig2000lof}; (iii) Isolation Forest (IF), a tree-based method
that isolates anomalies via random partitioning
\citep{liu2008isolation}; and (iv) a centroid-based score defined as the
Euclidean distance to the global mean of the embedding. Performance is evaluated using the area under the precision--recall curve (AUPRC), which is more informative in imbalanced anomaly detection settings. 
All metrics are averaged over 5 random seeds to account for stochasticity in the embedding procedures. We compare DR-SNE against standard dimensionality reduction methods, including \textit{t}-SNE, UMAP, and PaCMAP, as well as density-aware baselines such as DensMAP and DenSNE. 
For each method, we perform a grid search over key hyperparameters (two per method), covering a broad range of configurations. 
Full hyperparameter grids for each method can be found in Appendix~\ref{sec:anomaly_detection}. This evaluation framework mitigates some confounding factors by applying identical anomaly scoring functions across all embeddings. While downstream detectors depend on both geometry and density, this setup allows us to probe how differences in local density structure induced by the embedding affect anomaly detection performance.

Results are summarized in Table~\ref{tab:anomaly_table_three_main}. The left
panel reports Oracle-best performance, with configurations selected
independently for each method and detector, while the right panel reports
performance under Oracle-IF selection, with a single configuration per method
selected by maximizing Isolation Forest (IF) AUPRC. Additional results are
provided in Appendix~\ref{sec:anomaly_detection}.

Across datasets, performance is strongly detector- and dataset-dependent. On
\textbf{CIFAR}, DR-SNE is competitive but not dominant: DenSNE achieves the
highest kNN score, UMAP the highest LOF score, and PaCMAP the highest IF and
centroid scores under Oracle-best selection. On \textbf{Thyroid}, DR-SNE
achieves the highest kNN score and ties for the highest IF score under
Oracle-best selection, while also achieving the highest centroid score.
DensMAP performs best for LOF. Under Oracle-IF selection, HIGH-D achieves the
highest kNN score, DensMAP the highest LOF score, and DR-SNE ties for the
highest IF score and achieves the highest centroid score. On
\textbf{Shuttle}, PaCMAP performs substantially better for IF and centroid
scoring, whereas DR-SNE achieves the highest kNN score and is close to
DensMAP for LOF under Oracle-best selection. Under Oracle-IF selection,
HIGH-D achieves the highest kNN and LOF scores, while PaCMAP remains strongest
for IF and centroid scoring.

Overall, these results suggest that preservation of relative local concentration can provide useful information in some anomaly-detection settings, although the benefit is strongly dataset- and detector-dependent. No embedding method consistently dominates across the evaluated detectors. At the same time, geometry-focused methods may be preferable for detectors such as IF or centroid scoring that benefit from strong global separation. 

To further examine the role of concentration preservation in downstream
performance, we additionally vary the DR-SNE regularization strength
$\lambda$ and evaluate anomaly-detection AUPRC across detectors. The results
reveal two broad empirical regimes: on some datasets, small non-zero values
of $\lambda$ provide the strongest downstream performance, whereas on others,
most notably Thyroid, stronger concentration regularization remains
beneficial for several detectors. This analysis further illustrates that the
downstream utility of concentration preservation is both dataset- and
detector-dependent. Full results are provided in
Appendix~\ref{sec:lambda_anomaly}.

DR-SNE has $\mathcal{O}(n^2)$ time complexity per iteration due to dense
pairwise interactions, similar to standard t-SNE. Consequently, the current
implementation has relatively high computational cost and runtime,
particularly as the number of samples increases. While this limits
scalability, DR-SNE could in principle be combined with existing acceleration
techniques, such as Barnes--Hut or interpolation-based methods, although such
extensions are not implemented or evaluated in this work. A full complexity
and empirical runtime analysis is provided in
Appendix~\ref{sec:time_complexity}.
\section{Discussion}

The results demonstrate that preserving local neighborhood structure does not necessarily preserve relative local concentration. Classical neighborhood-preserving
methods optimize affinity or geometric objectives without explicitly
requiring preservation of pointwise neighborhood scale. Consequently,
strong neighborhood fidelity can coexist with substantial changes in
relative local concentration. DR-SNE addresses this distinction by
encouraging observations that are relatively concentrated in the
original space to remain relatively concentrated in the embedding, and
comparatively dispersed observations to remain dispersed.

Importantly, the concentration statistic used by DR-SNE measures inverse
characteristic neighborhood scale rather than dimensionally calibrated
probability density or local volume. This distinction is supported by
the objective-independent evaluation using a conventional $k$-NN
volume-density estimator: method rankings differ between concentration
correlation and $k$-NN density correlation, indicating that the two
criteria capture related but non-equivalent properties. The strong
performance of DR-SNE should therefore be interpreted as preservation
of its targeted relative-concentration profile, not as universal
superiority in probability-density preservation.

The regularization strength $\lambda$ controls the balance between
neighborhood fidelity and concentration preservation. Increasing
$\lambda$ generally improves preservation of the targeted concentration
profile, while sufficiently strong regularization can reduce
neighborhood-based fidelity. However, small non-zero values can improve
concentration preservation with little or no loss of local structure,
showing that the objectives are not necessarily strongly antagonistic
at moderate regularization strengths. The choice of neighborhood size
$k$ similarly determines the spatial scale at which concentration is
measured, so appropriate settings may depend on the data and intended
application.

These distinctions also clarify the relationship between DR-SNE and
density-aware methods such as DensMAP and DenSNE. The methods impose
different constraints on local scale or distributional structure and
should not be regarded as interchangeable implementations of a single
density-preservation objective. The controlled synthetic experiments,
where variation in local concentration is known by construction,
complement the real-world benchmarks by making these differences more
explicit.

The anomaly-detection experiments further suggest that preserving
relative local concentration can be useful when local crowding carries
task-relevant information, but the benefit is dataset- and
detector-dependent. DR-SNE is competitive and improves performance in
several settings, particularly for neighborhood-based detection, but
does not uniformly outperform other embeddings or the original
high-dimensional representation. These results therefore support
relative local concentration as a potentially useful downstream signal,
rather than a universally preferable embedding criterion.

Finally, the current implementation retains the $\mathcal{O}(n^2)$
pairwise cost of standard t-SNE and is therefore less scalable than
accelerated alternatives such as Barnes--Hut-based methods. Despite this
limitation, DR-SNE provides a simple mechanism for explicitly controlling
relative local concentration while retaining the SNE neighborhood
objective. More broadly, the differences between concentration-based and
conventional density-based evaluations emphasize the importance of
specifying precisely which notion of local-scale preservation is being
optimized and evaluated in dimensionality reduction.

\section{Conclusion}

We introduced DR-SNE, an extension of stochastic neighbor embedding that explicitly preserves relative local concentration by aligning normalized neighborhood-scale profiles between the original and embedding spaces. This complements the neighborhood-affinity objective of SNE by controlling variation in local scale across observations. Importantly, this concentration statistic is distinct from dimensionally calibrated probability density, as confirmed by independent $k$-NN volume-density evaluation.

Across diverse datasets, DR-SNE strongly preserves its targeted concentration profile while maintaining competitive neighborhood fidelity, with $\lambda$ controlling the trade-off between these properties. The results establish relative local concentration as a distinct embedding property that can be explicitly controlled alongside neighborhood structure.

\clearpage
\bibliographystyle{apalike}
\bibliography{references}

@article{maaten2008tsne,
  title   = {Visualizing Data using {t-SNE}},
  author  = {van der Maaten, Laurens and Hinton, Geoffrey},
  journal = {Journal of Machine Learning Research},
  volume  = {9},
  pages   = {2579--2605},
  year    = {2008}
}

@article{narayan2021density,
  title   = {Density-preserving data visualization unveils dynamic patterns of single-cell transcriptomic variability},
  author  = {Narayan, Ashwin and Berger, Bonnie and Cho, Hyunghoon},
  journal = {Nature Biotechnology},
  volume  = {39},
  pages   = {765--774},
  year    = {2021}
}

@book{jolliffe2002pca,
  title     = {Principal Component Analysis},
  author    = {Jolliffe, Ian T.},
  edition   = {2},
  publisher = {Springer},
  year      = {2002}
}

@article{tenenbaum2000isomap,
  title={A global geometric framework for nonlinear dimensionality reduction},
  author={Tenenbaum, Joshua B and de Silva, Vin and Langford, John C},
  journal={Science},
  volume={290},
  number={5500},
  pages={2319--2323},
  year={2000}
}

@article{roweis2000lle,
  title={Nonlinear dimensionality reduction by locally linear embedding},
  author={Roweis, Sam T. and Saul, Lawrence K.},
  journal={Science},
  volume={290},
  number={5500},
  pages={2323--2326},
  year={2000}
}

@article{wattenberg2016tsne,
  title   = {How to Use {t-SNE} Effectively},
  author  = {Wattenberg, Martin and Vi{\'e}gas, Fernanda and Johnson, Ian},
  journal = {Distill},
  year    = {2016},
  doi     = {10.23915/distill.00002}
}

@article{xiao2017fashion,
  title={Fashion-MNIST: a Novel Image Dataset for Benchmarking Machine Learning Algorithms},
  author={Xiao, Han and Rasul, Kashif and Vollgraf, Roland},
  journal={arXiv preprint arXiv:1708.07747},
  year={2017}
}

@article{pedregosa2011scikit,
  title   = {Scikit-learn: Machine Learning in Python},
  author  = {Pedregosa, Fabian and Varoquaux, Ga{\"e}l and Gramfort, Alexandre and Michel, Vincent and Thirion, Bertrand and Grisel, Olivier and Blondel, Mathieu and Prettenhofer, Peter and Weiss, Ron and Dubourg, Vincent and Vanderplas, Jake and Passos, Alexandre and Cournapeau, David and Brucher, Matthieu and Perrot, Matthieu and Duchesnay, {\'E}douard},
  journal = {Journal of Machine Learning Research},
  volume  = {12},
  pages   = {2825--2830},
  year    = {2011}
}

@article{zheng2017massively,
  title   = {Massively Parallel Digital Transcriptional Profiling of Single Cells},
  author  = {Zheng, Grace X. Y. and others},
  journal = {Nature Communications},
  volume  = {8},
  pages   = {14049},
  year    = {2017},
  doi     = {10.1038/ncomms14049}
}

@article{mcinnes2018umap,
  title={UMAP: Uniform Manifold Approximation and Projection for Dimension Reduction},
  author={McInnes, Leland and Healy, John and Melville, James},
  journal={arXiv preprint arXiv:1802.03426},
  year={2018}
}

@inproceedings{wang2021pacmap,
  title={PaCMAP: Pairwise Controlled Manifold Approximation Projection},
  author={Wang, Yingfan and Huang, Haifeng and Rudin, Cynthia and Shaposhnik, Yaron},
  booktitle={Proceedings of the 38th International Conference on Machine Learning (ICML)},
  pages={11109--11118},
  year={2021}
}

@inproceedings{amid2019trimap,
  title={TriMAP: Large-scale dimensionality reduction using triplets},
  author={Amid, Ehsan and Warmuth, Manfred K.},
  booktitle={Advances in Neural Information Processing Systems (NeurIPS)},
  volume={32},
  year={2019}
}

@article{van2014accelerating,
  title={Accelerating t-SNE using Tree-Based Algorithms},
  author={van der Maaten, Laurens},
  journal={Journal of Machine Learning Research},
  volume={15},
  number={93},
  pages={3221--3245},
  year={2014}
}

@article{linderman2019fast,
  title={Fast interpolation-based t-SNE for improved visualization of single-cell RNA-seq data},
  author={Linderman, George C. and Rachh, Manas and Hoskins, Jeremy G. and Steinerberger, Stefan and Kluger, Yuval},
  journal={Nature Methods},
  volume={16},
  number={3},
  pages={243--245},
  year={2019}
}

@article{loftsgaarden1965nonparametric,
  title={A nonparametric estimate of a multivariate density function},
  author={Loftsgaarden, Dean O. and Quesenberry, Charles P.},
  journal={The Annals of Mathematical Statistics},
  volume={36},
  number={3},
  pages={1049--1051},
  year={1965}
}

@inproceedings{venna2001neighborhood,
  title     = {Neighborhood Preservation in Nonlinear Projection Methods: An Experimental Study},
  author    = {Venna, Jarkko and Kaski, Samuel},
  booktitle = {Artificial Neural Networks---ICANN 2001},
  pages     = {485--491},
  publisher = {Springer},
  year      = {2001},
  doi       = {10.1007/3-540-44668-0_68}
}

@article{lecun1998mnist,
  title={Gradient-based learning applied to document recognition},
  author={LeCun, Yann and Bottou, L{\'e}on and Bengio, Yoshua and Haffner, Patrick},
  journal={Proceedings of the IEEE},
  volume={86},
  number={11},
  pages={2278--2324},
  year={1998}
}

@book{scott2015multivariate,
  title     = {Multivariate Density Estimation: Theory, Practice, and Visualization},
  author    = {Scott, David W.},
  edition   = {2},
  publisher = {Wiley},
  year      = {2015}
}

@techreport{krizhevsky2009learning,
  author       = {Krizhevsky, Alex},
  title        = {Learning Multiple Layers of Features from Tiny Images},
  institution  = {University of Toronto},
  year         = {2009}
}

@article{campos2016evaluation,
  title={On the Evaluation of Unsupervised Outlier Detection: Measures, Datasets, and an Empirical Study},
  author={Campos, Guilherme O. and Zimek, Arthur and Sander, J{\"o}rg and Campello, Ricardo J. G. B. and Micenkov{\'a}, Barbora and Schubert, Erich and Assent, Ira and Houle, Michael E.},
  journal={Data Mining and Knowledge Discovery},
  volume={30},
  number={4},
  pages={891--927},
  year={2016},
  doi={10.1007/s10618-015-0444-8}
}

@article{goldstein2016comparative,
  title={A Comparative Evaluation of Unsupervised Anomaly Detection Algorithms for Multivariate Data},
  author={Goldstein, Markus and Uchida, Seiichi},
  journal={PLOS ONE},
  volume={11},
  number={4},
  pages={e0152173},
  year={2016},
  doi={10.1371/journal.pone.0152173}
}

@inproceedings{ramaswamy2000efficient,
  title={Efficient Algorithms for Mining Outliers from Large Data Sets},
  author={Ramaswamy, Sridhar and Rastogi, Rajeev and Shim, Kyuseok},
  booktitle={Proceedings of the 2000 ACM SIGMOD International Conference
             on Management of Data},
  pages={427--438},
  year={2000},
  doi={10.1145/335191.335437}
}

@inproceedings{breunig2000lof,
  title={LOF: Identifying Density-Based Local Outliers},
  author={Breunig, Markus M. and Kriegel, Hans-Peter and Ng, Raymond T.
          and Sander, J{\"o}rg},
  booktitle={Proceedings of the 2000 ACM SIGMOD International Conference
             on Management of Data},
  pages={93--104},
  year={2000},
  doi={10.1145/342009.335388}
}

@inproceedings{liu2008isolation,
  title={Isolation Forest},
  author={Liu, Fei Tony and Ting, Kai Ming and Zhou, Zhi-Hua},
  booktitle={2008 Eighth IEEE International Conference on Data Mining},
  pages={413--422},
  year={2008},
  publisher={IEEE},
  doi={10.1109/ICDM.2008.17}
}

@misc{uci_thyroid,
  author    = {Quinlan, Ross},
  title     = {Thyroid Disease},
  year      = {1986},
  publisher = {UCI Machine Learning Repository},
  doi       = {10.24432/C5D010}
}

@misc{uci_shuttle,
  author    = {{UCI Machine Learning Repository}},
  title     = {Statlog (Shuttle)},
  year      = {1993},
  publisher = {UCI Machine Learning Repository},
  doi       = {10.24432/C5WS31}
}

\onecolumn
\clearpage
\appendix
\addcontentsline{toc}{section}{Appendix}
{
\hypersetup{linkcolor=black}
\tableofcontents
}
\newpage
\section{Theoretical Analysis of Relative Local Concentration}
\label{app:theory}

We analyze the concentration regularizer in DR-SNE and clarify the property it preserves. Because the regularizer operates on normalized inverse-neighborhood scales, it preserves relative variation in local neighborhood compactness across observations, rather than dimensionally calibrated probability density or local volume. We first formalize the relative local-concentration statistic and its normalization, then examine its relationship to conventional density estimation and the stochastic neighbor embedding objective. This analysis establishes the invariance properties of the proposed regularizer and clarifies what notion of local concentration DR-SNE preserves.

\subsection{Relative Local Concentration}

Let $X=\{x_i\}_{i=1}^n\subset\mathbb{R}^D$ denote the original data. For observation $x_i$, define the mean distance to its $k$ nearest
neighbors as
\begin{equation}
r_i^{(X)}
=
\frac{1}{k}
\sum_{j\in\mathcal{N}_k(i)}
\|x_i-x_j\|,
\end{equation}
where $\mathcal{N}_k(i)$ denotes the set of $k$ nearest neighbors of
$x_i$ in the original space.
We define the corresponding local concentration statistic as
\begin{equation}
c_i^{(X)}
=
\frac{1}{r_i^{(X)}+\epsilon},
\label{eq:local_concentration_appendix}
\end{equation}
where $\epsilon>0$ is a small constant included for numerical
stability. An analogous statistic $c_i^{(Z)}$ is computed from
neighborhood distances in the embedding space.

The statistic $c_i$ measures inverse characteristic neighborhood
scale: observations whose neighbors lie at smaller average distances
have larger concentration values, whereas observations with more
spatially dispersed neighborhoods have smaller values. Importantly, $c_i$ is not a dimensionally calibrated estimator of
probability density. Classical $k$-nearest-neighbor density estimation relates probability
density to the volume occupied by a neighborhood
~\citep{loftsgaarden1965nonparametric,scott2015multivariate}. Ignoring
constants, an $m$-dimensional $k$-NN volume-density estimator has the
form
\begin{equation}
\hat p_i
\propto
r_{i,k}^{-m},
\label{eq:knn_volume_density}
\end{equation}
where $m$ denotes the dimensionality of the corresponding space
($m=D$ in the original space and $m=d$ in the embedding), and
$r_{i,k}$ is the distance to the $k$th nearest neighbor.
Consequently, conventional density estimates depend explicitly on the
dimensionality of the space. By contrast, Eq.~\eqref{eq:local_concentration_appendix} uses an inverse length scale
rather than an inverse neighborhood volume. This distinction is particularly important in dimensionality
reduction, where the original and embedding spaces generally have
different dimensionalities. DR-SNE does not attempt to equate
dimensionally calibrated probability-density values across these
spaces. Instead, it seeks to preserve how each observation's
characteristic neighborhood scale compares with those of other
observations in the same space. To remove arbitrary global scale
differences, we normalize the concentration statistics to unit mean:
\begin{equation}
\tilde c_i^{(X)}
=
\frac{
c_i^{(X)}
}{
\frac{1}{n}\sum_{\ell=1}^{n}c_\ell^{(X)}
},
\qquad
\tilde c_i^{(Z)}
=
\frac{
c_i^{(Z)}
}{
\frac{1}{n}\sum_{\ell=1}^{n}c_\ell^{(Z)}
}.
\label{eq:normalized_concentration_appendix}
\end{equation}

The resulting quantities describe dataset-relative concentration
profiles rather than absolute density values.

\subsection{Concentration Regularization}

DR-SNE penalizes discrepancies between the normalized concentration
profiles of the original and embedding spaces:
\begin{equation}
\mathcal{L}_{\mathrm{conc}}
=
\frac{1}{n}
\sum_{i=1}^{n}
\left(
\log\tilde c_i^{(X)}
-
\log\tilde c_i^{(Z)}
\right)^2.
\label{eq:concentration_loss_appendix}
\end{equation}

Substituting Eq.~\eqref{eq:normalized_concentration_appendix} gives
\begin{equation}
\log\tilde c_i^{(X)}
-
\log\tilde c_i^{(Z)}
=
\log c_i^{(X)}
-
\log c_i^{(Z)}
+
C,
\end{equation}
where
\begin{equation}
C(Z)
=
\log
\left(
\frac{
\frac{1}{n}\sum_j c_j^{(Z)}
}{
\frac{1}{n}\sum_j c_j^{(X)}
}
\right)
\end{equation}
is a global offset shared across observations. Although independent of
$i$, it varies with the embedding through $c_j^{(Z)}$. Moreover,
\begin{equation}
\log c_i^{(X)}-\log c_i^{(Z)}
=
\log\left(
\frac{c_i^{(X)}}{c_i^{(Z)}}
\right),
\end{equation}
so the log representation expresses concentration discrepancies as
multiplicative ratios rather than absolute differences.

This objective is distinct from direct local-volume preservation.
Consider an idealized smooth, locally invertible mapping between spaces
of the same intrinsic dimension $m$. By the change-of-variables formula,
\begin{equation}
p_Z(f(x))
=
\frac{p_X(x)}
{J_{\mathcal M}f(x)},
\end{equation}
where $J_{\mathcal M}f(x)$ is the local volume expansion factor. Hence,
\begin{equation}
\log p_X(x)-\log p_Z(f(x))
=
\log J_{\mathcal M}f(x).
\end{equation}
Under these restrictive assumptions, matching dimensionally calibrated
probability densities can therefore be related to local volume
distortion. DR-SNE does not directly optimize this quantity. Its concentration
statistic scales with inverse characteristic neighborhood length rather
than inverse $m$-dimensional neighborhood volume. Accordingly,
$\mathcal{L}_{\mathrm{conc}}$ regularizes variation in relative local
neighborhood scale rather than directly minimizing Jacobian or
probability-density distortion. These notions can nevertheless be empirically related: preserving
relative local concentration may also preserve structure measured by
conventional density estimators, but the two properties need not
coincide. 

\paragraph{Interpretation.}
Minimizing $\mathcal{L}_{\mathrm{conc}}$ encourages observations with
relatively compact neighborhoods in the original space to remain
relatively compact in the embedding, and dispersed neighborhoods to
remain comparatively dispersed. The regularizer therefore complements
neighborhood-affinity preservation by explicitly constraining relative
variation in local neighborhood scale.

\subsection{Choice of the Inverse Neighborhood Scale}

The inverse neighborhood scale in
Eq.~\ref{eq:local_concentration_appendix} corresponds to $\alpha=1$
in the more general family
\begin{equation}
c_i^{(\alpha)}
=
r_i^{-\alpha},
\qquad \alpha>0,
\end{equation}
where $r_i$ denotes the characteristic neighborhood scale. Any
$\alpha>0$ preserves the ordering of local compactness,
\begin{equation}
r_i < r_j
\quad\Longleftrightarrow\quad
c_i^{(\alpha)} > c_j^{(\alpha)}.
\end{equation}

The exponent controls the emphasis placed on differences in
neighborhood scale. Ignoring the numerical-stability constant
$\epsilon$, mean normalization gives
\begin{equation}
\tilde{c}_i^{(\alpha)}
=
\frac{r_i^{-\alpha}}
{\frac{1}{n}\sum_j r_j^{-\alpha}},
\end{equation}
and therefore
\begin{equation}
\log \tilde{c}_i^{(\alpha)}
=
-\alpha\log r_i
-
\log\left(
\frac{1}{n}\sum_j r_j^{-\alpha}
\right).
\end{equation}
Thus, $\alpha$ controls the sensitivity of the regularizer to
differences in neighborhood scale. We use $\alpha=1$, for which
concentration ratios have the direct interpretation
\begin{equation}
\frac{c_i}{c_j}
=
\frac{r_j}{r_i}.
\end{equation}
This choice therefore preserves an inverse-length interpretation without
introducing an additional shape parameter. For comparison, an
$m$-dimensional volume-based density estimator has characteristic
scaling $r^{-m}$, with an exponent that depends explicitly on
dimensionality. DR-SNE instead applies the same inverse-length
transformation in the original and embedding spaces.

\subsection{Invariance Properties}

The invariance properties hold for the general family
$c_i^{(\alpha)}=r_i^{-\alpha}$, $\alpha>0$. Under a global rescaling
$z_i'=a z_i$, $a>0$, neighborhood scales satisfy
\begin{equation}
r_i^{(Z')}=a r_i^{(Z)},
\qquad
c_i^{(Z')}=a^{-\alpha}c_i^{(Z)}.
\end{equation}
Ignoring the numerical-stability constant $\epsilon$, the common factor
$a^{-\alpha}$ cancels under mean normalization:
\begin{equation}
\tilde c_i^{(Z')}
=
\frac{a^{-\alpha}c_i^{(Z)}}
{a^{-\alpha}\frac{1}{n}\sum_j c_j^{(Z)}}
=
\tilde c_i^{(Z)}.
\end{equation}
Thus, $\mathcal{L}_{\mathrm{conc}}$ is invariant to global rescaling
for any $\alpha>0$ in the $\epsilon=0$ limit.

The regularizer is also translation invariant, since for
$z_i'=z_i+b$,
\begin{equation}
\|z_i'-z_j'\|=\|z_i-z_j\|.
\end{equation}
Consequently, it constrains relative variation in neighborhood scale
rather than global position or absolute scale.

\subsection{Pointwise versus Correlation-Based Regularization}

DR-SNE matches normalized local-concentration profiles pointwise.
In contrast, correlation-based objectives constrain the association
between local-scale statistics. Perfect correlation in log-space is
compatible with
\begin{equation}
\log R_i^{(Z)}
=
\beta\log R_i^{(X)}+b,
\qquad \beta>0,
\end{equation}
or equivalently,
\begin{equation}
R_i^{(Z)}
=
A\left(R_i^{(X)}\right)^\beta.
\end{equation}
Thus, perfect correlation can be maintained when relative scale
differences are amplified ($\beta>1$) or compressed
($0<\beta<1$). By contrast, zero pointwise DR-SNE loss requires
preservation of relative neighborhood-scale ratios up to a common
global scale factor, as shown above. The two objectives therefore
encode different notions of local-scale preservation.
\clearpage
\section{Visualization}
\subsection{Implementation details}
\label{sec:implementation_details}
All methods are implemented in Python using standard libraries and official reference implementations. We use \texttt{scikit-learn} for t-SNE, \texttt{umap-learn} for UMAP and DensMAP, and the official \texttt{pacmap} package for PaCMAP. DR-SNE is implemented in PyTorch using a custom implementation optimized with Adam for 1000 iterations with Gaussian initialization. DenSNE is implemented using the official \texttt{densvis} package. For the main visualization comparison, one key hyperparameter per method is tuned, while all other parameters are fixed.

The following hyperparameter grids are used:

\textbf{DR-SNE:}
\[
\lambda \in \{10^{-4}, 2.5\!\times\!10^{-4}, 5\!\times\!10^{-4}, 10^{-3}, 2.5\!\times\!10^{-3}, 5\!\times\!10^{-3}, 10^{-2}, 2.5\!\times\!10^{-2}, 5\!\times\!10^{-2}, 10^{-1}\}
\]
Other parameters: iterations $=1000$, density neighborhood size $k=300$, Adam optimizer, Gaussian initialization.

\textbf{DenSNE:}
\[
\lambda \in \{10^{-3}, 2.5\!\times\!10^{-3}, 5\!\times\!10^{-3}, 10^{-2}, 2.5\!\times\!10^{-2}, 5\!\times\!10^{-2}, 10^{-1}, 2.5\!\times\!10^{-1}, 5\!\times\!10^{-1}, 1.0\}
\]
Other parameters: perplexity $=30$, iterations $=800$, $\theta=0.5$, $\texttt{dens\_frac}=0.5$, $\texttt{use\_pca=False}$, Gaussian initialization ($10^{-4}$), $\texttt{final\_dens=True}$.

\textbf{t-SNE:}
\[
\text{perplexity} \in \{3, 5, 10, 20, 30, 40, 50, 75, 100, 150\}
\]
Other parameters: random initialization, default learning rate and iterations (scikit-learn).

\textbf{UMAP:}
\[
n_{\text{neighbors}} \in \{3, 5, 10, 20, 30, 40, 50, 75, 100, 150\}
\]
Other parameters: $\text{min\_dist}=0.1$, embedding dimension $=2$.

\textbf{PaCMAP:}
\[
n_{\text{neighbors}} \in \{3, 5, 10, 20, 30, 40, 50, 75, 100, 150\}
\]
Other parameters: $\texttt{MN\_ratio}=0.5$, $\texttt{FP\_ratio}=2.0$, embedding dimension $=2$.

\textbf{DensMAP:}
\[
\lambda \in \{0.01, 0.025, 0.05, 0.1, 0.25, 0.5, 1.0, 2.5, 5.0, 10.0\}
\]
Other parameters: $n_{\text{neighbors}}=15$, $\text{min\_dist}=0.1$, $\texttt{densmap=True}$, $\texttt{dens\_frac}=0.3$.

All experiments are repeated over 5 random seeds $\{0,1,2,3,4\}$.
\subsection{Synthetic datasets}
\label{sec:synthetic_datasets}

To provide controlled settings in which relative local concentration is
known by construction, we introduce two three-dimensional synthetic
benchmarks: Controlled Clusters and Concentration-Gradient Helix.

\paragraph{Controlled Clusters.}
The Controlled Clusters dataset consists of three approximately equally
sized isotropic Gaussian populations,
\begin{equation}
    \mathbf{x}_{ij}
    \sim
    \mathcal{N}(\boldsymbol{\mu}_i,\sigma_i^2 I_3),
\end{equation}
with centers
$\boldsymbol{\mu}_1=(-6,0,0)$,
$\boldsymbol{\mu}_2=(0,6,0)$, and
$\boldsymbol{\mu}_3=(6,0,0)$, and scales
$(\sigma_1,\sigma_2,\sigma_3)=(0.20,0.50,1.00)$.
We generate $N=5000$ observations without standardization, producing
three populations with deliberately different characteristic scales.

\paragraph{Concentration-Gradient Helix.}
The Concentration-Gradient Helix tests continuously varying
concentration along a connected nonlinear manifold. We sample
\begin{equation}
    p(u)=\frac{\alpha e^{-\alpha u}}{1-e^{-\alpha}},
    \qquad u\in[0,1],\quad \alpha=3,
\end{equation}
and map the samples to
\begin{equation}
    t = 2\pi n_{\mathrm{turns}}u,
\end{equation}
\begin{equation}
    \mathbf{x}
    =
    (\cos t,\sin t,pt)^\top
    + \boldsymbol{\epsilon},
\end{equation}
where
\begin{equation}
    \boldsymbol{\epsilon}
    \sim
    \mathcal{N}(0,0.025^2 I_3).
\end{equation}
Here $n_{\mathrm{turns}}=4$ and $p=0.35$. We generate $N=5000$
observations. Since $p(u)$ decreases monotonically, sampling
concentration decreases continuously along the helix. The ten colors
shown in Figure~\ref{fig:main_real_world} correspond to intervals of
$u$ for visualization only and are not used by the embedding methods.
\subsection{Results}
\begin{figure*}[!tb]
\centering
\includegraphics[width=0.8\textwidth]{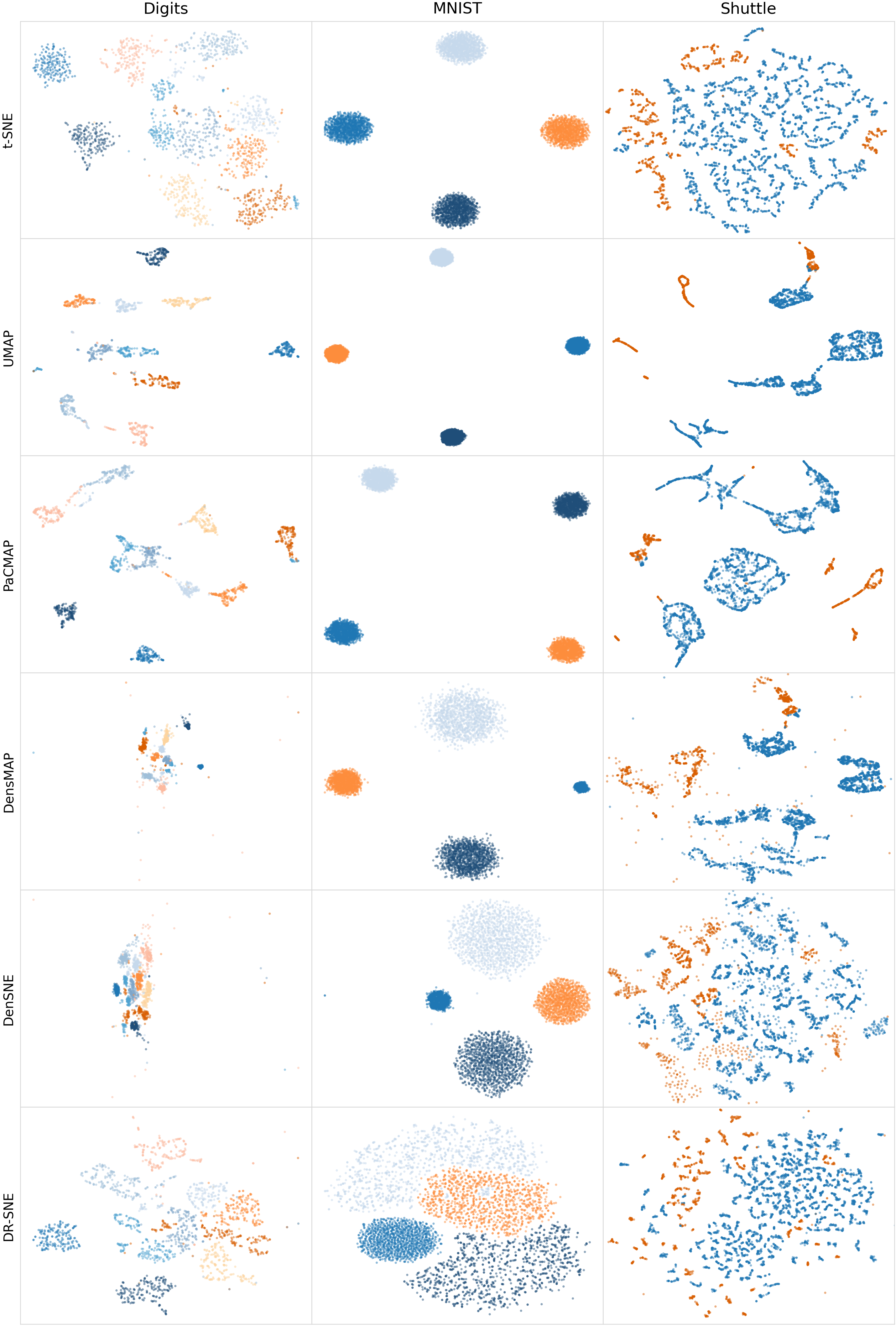}
\caption{
Comparison of dimensionality reduction methods on additional benchmark datasets.
Rows correspond to different methods (t-SNE, UMAP, PaCMAP, DensMAP, DenSNE, and DR-SNE), and columns correspond to Digits, MNIST, and Shuttle.
}
\label{fig:appendix_comparison}
\end{figure*}

Figure~\ref{fig:appendix_comparison} extends the qualitative comparison in Figure~\ref{fig:main_real_world} to Digits, MNIST, and Shuttle. On \textbf{Digits}, t-SNE produces a relatively diffuse arrangement with several partially separated groups, whereas UMAP and PaCMAP compress the data into smaller, more fragmented structures. DensMAP and DenSNE reorganize the embedding more strongly, while DR-SNE retains a broader multi-group layout together with pronounced differences in local dispersion. DensMAP and DenSNE substantially reorganize these structures, whereas DR-SNE retains the broad multi-cluster organization while expressing pronounced differences in local concentration across groups. On \textbf{MNIST}, the contrast is especially clear because the displayed groups are well separated: t-SNE, UMAP, and PaCMAP represent them as compact clusters of relatively similar scale, while DensMAP and DenSNE introduce stronger differences in their dispersion. DR-SNE further accentuates these relative differences, producing pronounced variation between compact and diffuse populations, although with some reduction in cluster separation. On \textbf{Shuttle}, the methods produce markedly different organizations of the data. t-SNE, UMAP, and PaCMAP form multiple compact or elongated structures, while DensMAP and DenSNE produce more strongly redistributed configurations with substantial variation in local scale. DR-SNE yields a distinct arrangement with pronounced differences between locally compact and dispersed regions, accompanied by substantial reorganization of the global structure. Overall, these additional examples are consistent with the main-text observations: DR-SNE produces pronounced differences in local concentration, although this can be accompanied by changes in cluster separation and global organization that vary across datassets.

\FloatBarrier
\label{sec:visualization}
\clearpage

\section{Additional Ablation Studies}
\label{sec:ablation_studies}
\begin{figure}[H]
\centering
\includegraphics[width=\linewidth]{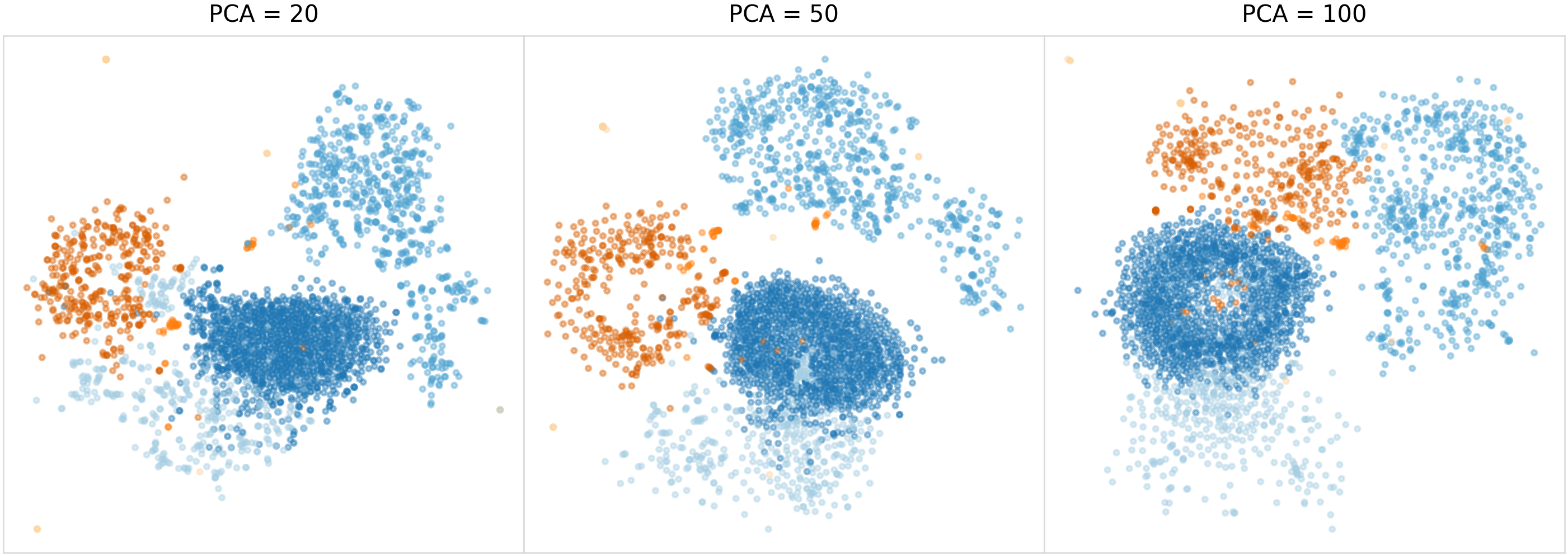}
\caption{
Effect of PCA dimensionality on the embedding with fixed concentration
regularization ($\lambda = 0.01$) and neighborhood size ($k = 50$).
From left to right, the number of retained principal components increases
(20, 50, 100).
}
\label{fig:pca_ablation}
\end{figure}

Here we present additional ablation studies on PCA dimensionality and
neighborhood size on the PBMC dataset. We first analyze the effect of PCA
dimensionality on embedding quality under fixed regularization strength
$\lambda = 0.01$ (Table~\ref{tab:pca_ablation_pbmc}). The results reveal a
clear trade-off between local neighborhood preservation and global
pairwise-distance fidelity. Lower-dimensional representations
(PCA = 10--20) achieve the highest trustworthiness and continuity, indicating
stronger preservation of local neighborhoods. As the number of principal
components increases, both metrics generally decline, while stress decreases
substantially, indicating improved preservation of global pairwise geometry.

In contrast, concentration correlation remains high across all evaluated PCA
dimensionalities, ranging from $0.968$ to $0.988$. This suggests that
preservation of the targeted relative local-concentration profile is
comparatively robust to PCA dimensionality, even as neighborhood and global
geometric properties change substantially. The transition is gradual rather
than abrupt, indicating a continuous trade-off between local neighborhood
preservation and global geometric fidelity rather than a narrow optimal PCA
regime.

% --- BOTTOM TABLE ---
\begin{table}[!htbp]
\centering
\scriptsize
\setlength{\tabcolsep}{5pt}
\renewcommand{\arraystretch}{1.15}
\caption{Ablation over PCA dimensionality with fixed $k=50$ and
$\lambda=0.01$ on the PBMC dataset. Results are mean $\pm$
standard deviation over 3 runs.}
\label{tab:pca_ablation_pbmc}
\vspace{5pt}
\begin{tabular}{lccccc}
\toprule
PCA 
& Trustworthiness 
& Continuity 
& Concentration Corr. 
& Silhouette 
& Stress \\
\midrule
10  & $\mathbf{0.910\pm0.009}$ & $\mathbf{0.948\pm0.004}$ & $0.983\pm0.001$ & $\mathbf{0.144\pm0.044}$ & $3.253\pm0.198$ \\
20  & $0.873\pm0.009$ & $0.901\pm0.009$ & $0.973\pm0.001$ & $0.128\pm0.043$ & $2.502\pm0.078$ \\
40  & $0.819\pm0.006$ & $0.863\pm0.007$ & $0.968\pm0.001$ & $0.070\pm0.017$ & $1.831\pm0.027$ \\
60  & $0.760\pm0.004$ & $0.780\pm0.025$ & $0.969\pm0.001$ & $0.060\pm0.032$ & $1.349\pm0.053$ \\
80  & $0.716\pm0.003$ & $0.671\pm0.001$ & $0.977\pm0.001$ & $-0.046\pm0.012$ & $1.081\pm0.005$ \\
100 & $0.711\pm0.004$ & $0.651\pm0.005$ & $0.984\pm0.000$ & $-0.028\pm0.008$ & $0.899\pm0.030$ \\
120 & $0.707\pm0.004$ & $0.678\pm0.006$ & $0.987\pm0.000$ & $-0.025\pm0.003$ & $0.625\pm0.014$ \\
140 & $0.697\pm0.002$ & $0.692\pm0.007$ & $\mathbf{0.988\pm0.001}$ & $-0.010\pm0.002$ & $\mathbf{0.445\pm0.002}$ \\
\bottomrule
\end{tabular}
\end{table}

% =========================
% PAGE 2 (k ablation)
% =========================

% --- TOP FIGURE ---
\begin{figure}[t]
\centering
\includegraphics[width=\linewidth]{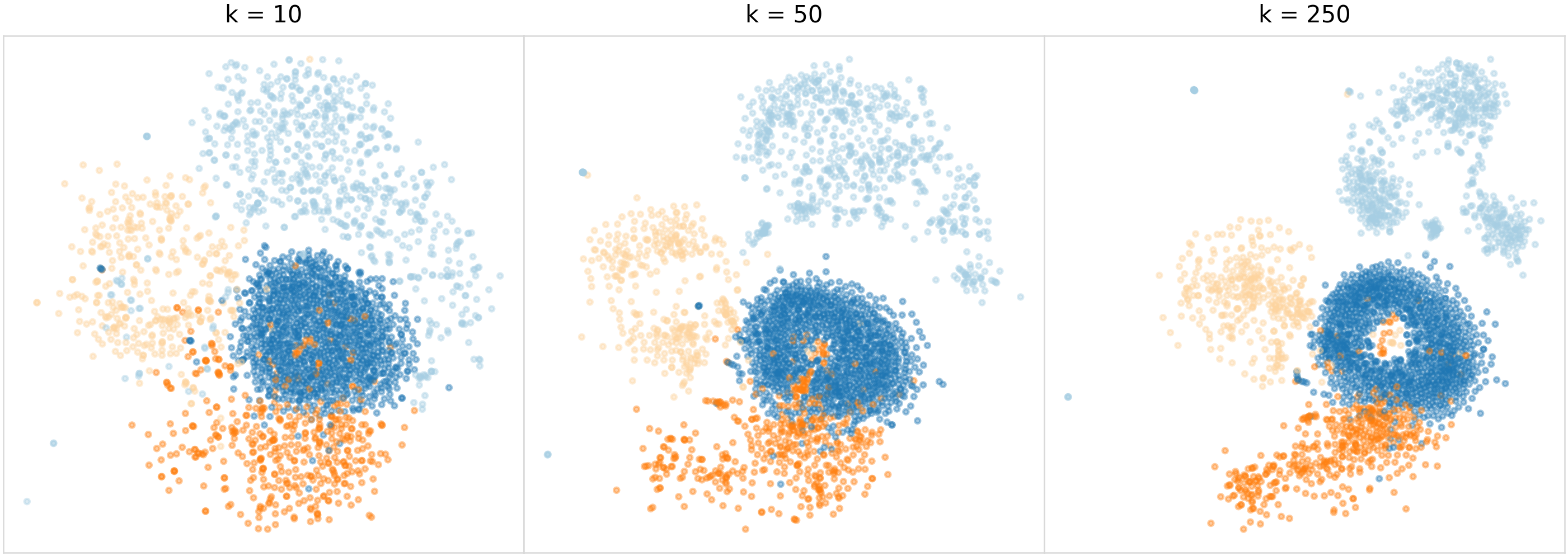}
\caption{
Effect of neighborhood size ($k$) on the DR-SNE embedding of the PBMC dataset.
We vary the neighborhood size
($k \in \{10, 50, 250\}$) while keeping all other parameters fixed
($\lambda = 0.01$, $PCA = 30$).
}
\label{fig:neighbors_ablation}
\end{figure}

We further analyze the effect of neighborhood size $k$ on embedding behavior
(Fig.~\ref{fig:neighbors_ablation}, Table~\ref{tab:k_ablation_pbmc}).
Increasing $k$ substantially improves neighborhood preservation:
trustworthiness increases from $0.720$ at $k=10$ to $0.905$ at $k=400$,
while continuity increases from $0.776$ to $0.922$. Silhouette scores show a
similar overall improvement. Stress decreases sharply up to $k=200$, where
it reaches its minimum, and then increases slightly for larger neighborhoods.
In contrast, concentration correlation remains consistently high across the
entire range ($0.974$--$0.989$), reaching its maximum at $k=200$.

These results indicate that neighborhood size has a pronounced effect on
local neighborhood preservation and global pairwise geometry, while
preservation of the targeted relative local-concentration profile is
comparatively insensitive to this parameter. Most of the improvement occurs
between $k=10$ and $k\approx150$--$200$; beyond this range, trustworthiness
and continuity largely saturate, while stress no longer improves.

Increasing $k$ also changes the spatial scale at which local concentration is
measured. Larger neighborhoods average over broader regions of the data and
are therefore less sensitive to fine-scale local variation, whereas smaller
values emphasize more localized structure. The results consequently suggest
that intermediate neighborhood sizes, around $k=150$--$200$ in this
experiment, provide a favorable balance between neighborhood fidelity,
global geometric preservation, and concentration alignment.

\vfill

% --- BOTTOM TABLE ---
\begin{table}[t]
\centering
\scriptsize
\setlength{\tabcolsep}{5pt}
\renewcommand{\arraystretch}{1.15}
\caption{Ablation over neighborhood size $k$ with fixed PCA dimensionality
of 30 and $\lambda=0.01$ on the PBMC dataset. Results are mean $\pm$
standard deviation over 3 runs.}
\label{tab:k_ablation_pbmc}
\begin{tabular}{lccccc}
\toprule
$k$ 
& Trustworthiness 
& Continuity 
& Concentration Corr. 
& Silhouette 
& Stress \\
\midrule
10  & $0.720\pm0.023$ & $0.776\pm0.022$ & $0.974\pm0.001$ & $-0.028\pm0.011$ & $2.183\pm0.041$ \\
50  & $0.833\pm0.020$ & $0.876\pm0.011$ & $0.974\pm0.002$ & $0.169\pm0.028$ & $1.786\pm0.063$ \\
100 & $0.873\pm0.007$ & $0.902\pm0.006$ & $0.982\pm0.001$ & $0.248\pm0.056$ & $1.013\pm0.048$ \\
150 & $0.895\pm0.001$ & $0.915\pm0.003$ & $0.987\pm0.001$ & $0.328\pm0.009$ & $0.459\pm0.034$ \\
200 & $0.898\pm0.001$ & $0.920\pm0.002$ & $\mathbf{0.989\pm0.000}$ & $0.326\pm0.024$ & $\mathbf{0.432\pm0.003}$ \\
250 & $0.898\pm0.005$ & $0.919\pm0.003$ & $0.983\pm0.009$ & $0.333\pm0.014$ & $0.467\pm0.035$ \\
300 & $0.901\pm0.002$ & $0.920\pm0.000$ & $0.983\pm0.004$ & $0.344\pm0.013$ & $0.505\pm0.005$ \\
350 & $0.903\pm0.003$ & $0.921\pm0.004$ & $0.988\pm0.001$ & $0.353\pm0.009$ & $0.532\pm0.015$ \\
400 & $\mathbf{0.905\pm0.002}$ & $\mathbf{0.922\pm0.002}$ & $0.988\pm0.002$ & $\mathbf{0.367\pm0.022}$ & $0.555\pm0.001$ \\
\bottomrule
\end{tabular}
\end{table}

\FloatBarrier
\clearpage
\section{Downstream tasks: Anomaly Detection}
\label{sec:anomaly_detection}

\paragraph{Experimental setup.}
All methods are evaluated under a unified pipeline: inputs are standardized, embeddings are computed in 2D, and anomaly scores are obtained using kNN, LOF, Isolation Forest (IF), and centroid-based scoring. Results in the main text and Appendix follow the same evaluation pipeline but are reported on different datasets.
Both \textbf{Best} (detector-specific tuning) and \textbf{Fair (IF)} selection protocols are considered. The same setup is applied consistently across all datasets and methods to ensure a controlled and comparable evaluation.

\paragraph{Datasets}

We evaluate on a combination of real-world and synthetic datasets designed to probe density-aware anomaly detection. \textbf{PBMC}~\citep{zheng2017massively} is a single-cell RNA-seq dataset where rare cell populations are treated as anomalies; cells are normalized, log-transformed, reduced via PCA, and clustered using Leiden, with the smallest clusters defining anomalies. \textbf{Fashion-MNIST}~\citep{xiao2017fashion} is used in a one-class anomaly detection setting, where a single class is treated as normal and all remaining classes are considered anomalous, providing a high-dimensional benchmark with semantic rather than density-based anomalies. \textbf{CIFAR-10}~\citep{krizhevsky2009learning} is included as a more complex natural image dataset, where we adopt the same one-class protocol, treating one class as inliers and the rest as anomalies, resulting in a challenging high-dimensional setting with less separable structure.
\textbf{Shuttle}~\citep{uci_shuttle} is a tabular dataset with a well-defined anomaly class, commonly used in anomaly detection benchmarks; it represents a density-driven regime where anomalies correspond to rare system states. Synthetic datasets explicitly control density variations. In the \textbf{density spiral dataset}, samples are generated along a spiral manifold $(x,y)=(t\cos t, t\sin t)$ with sinusoidal density modulation. Points are resampled according to this density, perturbed with small Gaussian noise, and embedded into higher-dimensional space via random linear projections followed by standardization. Anomalies are defined as samples below a density percentile threshold.

\begin{table}[!b]
\centering
\scriptsize
\setlength{\tabcolsep}{2pt}
\renewcommand{\arraystretch}{1.00}

\caption{
AUPRC (Area Under the Precision--Recall Curve; mean $\pm$ standard deviation) over 5 runs
for kNN (k-Nearest Neighbors), LOF (Local Outlier Factor), IF (Isolation
Forest), and Cent (Centroid-based detector).
Left: oracle-best performance, where the embedding hyperparameters are
selected independently for each method and detector by maximizing AUPRC
using the ground-truth anomaly labels.
Right: oracle-IF selection, where a single embedding configuration for
each method is selected by maximizing Isolation Forest AUPRC and is
subsequently evaluated with all four detectors.
Because both selection procedures use ground-truth anomaly labels,
these results should be interpreted as label-informed upper-bound
analyses rather than as fully unsupervised model-selection protocols.
Each dataset caption reports the number of samples ($n$) and anomaly
rate (a.r.).
}
\vspace{8pt}
% ===================== PBMC =====================
\textbf{PBMC}, $n = 2700$,  $a.r. = 0.05$
\vspace{2pt}
\resizebox{\textwidth}{!}{
\begin{tabular}{lcccc|cccc l}
\toprule
 & \multicolumn{4}{c}{Best} & \multicolumn{4}{c}{Fair (IF)} &  \\
\cmidrule(lr){2-5} \cmidrule(lr){6-9}
Method & kNN & LOF & IF & Cent & kNN & LOF & IF & Cent & Params \\
\midrule

\textbf{DR-SNE }  & .593 $\pm$ .037 & .500 $\pm$ .023 & \textbf{.632 $\pm$ .038} & \textbf{.648 $\pm$ .074}
        & .593 $\pm$ .037 & .392 $\pm$ .081 & \textbf{.632 $\pm$ .038} & \textbf{.625 $\pm$ .071} & $\lambda=2\times10^{-4}$, $k=80$ \\

DenSNE  & .712 $\pm$ .053 & .489 $\pm$ .073 & .606 $\pm$ .041 & .543 $\pm$ .089
        & \textbf{.624 $\pm$ .064} & .426 $\pm$ .103 & .606 $\pm$ .041 & .190 $\pm$ .104 & p=5, $\lambda$=0.1 \\

DensMAP & .708 $\pm$ .054 & \textbf{.595 $\pm$ .012} & .589 $\pm$ .061 & .559 $\pm$ .002
        & .589 $\pm$ .024 & .548 $\pm$ .039 & .589 $\pm$ .061 & .462 $\pm$ .112 & $\lambda$=0.2, f=0.3 \\

HIGH-D  & .543 $\pm$ .000 & .362 $\pm$ .000 & .257 $\pm$ .000 & .551 $\pm$ .000
        & .543 $\pm$ .000 & .362 $\pm$ .000 & .257 $\pm$ .000 & .551 $\pm$ .000 & -- \\

PaCMAP  & .726 $\pm$ .003 & .429 $\pm$ .095 & .600 $\pm$ .065 & .446 $\pm$ .272
        & .427 $\pm$ .117 & .429 $\pm$ .095 & .600 $\pm$ .065 & .303 $\pm$ .144 & k=5, FP=5 \\

UMAP    & \textbf{.732 $\pm$ .022} & .576 $\pm$ .114 & .406 $\pm$ .209 & .182 $\pm$ .174
        & .615 $\pm$ .064 & \textbf{.576 $\pm$ .114} & .406 $\pm$ .209 & .570 $\pm$ .116 & k=5, md=0.1 \\

t-SNE   & .712 $\pm$ .047 & .303 $\pm$ .101 & .599 $\pm$ .075 & .321 $\pm$ .196
        & .598 $\pm$ .078 & .103 $\pm$ .102 & .599 $\pm$ .075 & .521 $\pm$ .258 & p=100, lr=500 \\

\bottomrule
\vspace{6pt}
\end{tabular}
}

% ===================== Fashion =====================
\textbf{Fashion MNIST}, $n = 5000$, $a.r. = 0.1$
\vspace{2pt}

\resizebox{\textwidth}{!}{
\begin{tabular}{lcccc|cccc l}
\toprule
 & \multicolumn{4}{c}{Best} & \multicolumn{4}{c}{Fair (IF)} &  \\
\cmidrule(lr){2-5} \cmidrule(lr){6-9}
Method & kNN & LOF & IF & Cent & kNN & LOF & IF & Cent & Params \\
\midrule

\textbf{DR-SNE}  
& .368 $\pm$ .021 & .270 $\pm$ .024 & .513 $\pm$ .062 & .459 $\pm$ .024
& .290 $\pm$ .023 & .167 $\pm$ .014 & .513 $\pm$ .062 & .451 $\pm$ .040
& $\lambda$=2e-04, k=40 \\

DenSNE  
& .483 $\pm$ .036 & .199 $\pm$ .007 & .534 $\pm$ .043 & \textbf{.600 $\pm$ .065}
& .455 $\pm$ .021 & .140 $\pm$ .011 & .534 $\pm$ .043 & .569 $\pm$ .038
& p=100, $\lambda$=0.5 \\

DensMAP 
& \textbf{.506 $\pm$ .028} & .225 $\pm$ .026 & .574 $\pm$ .028 & .595 $\pm$ .018
&  \textbf{.479 $\pm$ .014} & .178 $\pm$ .006 & .574 $\pm$ .028 & .588 $\pm$ .018
& $\lambda$=0.1, f=0.3 \\

HIGH-D  
& .423 $\pm$ .000 & .114 $\pm$ .000 & .537 $\pm$ .000 & .459 $\pm$ .000
& .423 $\pm$ .000 & .114 $\pm$ .000 & .537 $\pm$ .000 & .459 $\pm$ .000
& -- \\

PaCMAP  
& .182 $\pm$ .022 & .215 $\pm$ .005 & .578 $\pm$ .010 & .533 $\pm$ .039
& .131 $\pm$ .014 & .167 $\pm$ .025 & .578 $\pm$ .010 & .477 $\pm$ .016
& k=8, FP=2 \\

UMAP    
& .379 $\pm$ .027 & \textbf{.313 $\pm$ .010} & \textbf{.602 $\pm$ .028} & .595 $\pm$ .020
& .132 $\pm$ .010 & \textbf{.194 $\pm$ .020} & \textbf{.602 $\pm$ .028} &  \textbf{.595 $\pm$ .020}
& k=8 \\

t-SNE   
& .254 $\pm$ .040 & .204 $\pm$ .005 & .398 $\pm$ .052 & .558 $\pm$ .015
& .128 $\pm$ .003 & .179 $\pm$ .005 & .398 $\pm$ .052 & .546 $\pm$ .007
& p=50 \\

\bottomrule
\end{tabular}
}
\vspace{6pt}

\label{tab:anomaly_auprc}
\end{table}
\paragraph{Grid Search} All methods are evaluated in 2D with inputs standardized prior to embedding (after dataset-specific preprocessing) and identical anomaly scoring pipelines (kNN, LOF, Isolation Forest, centroid distance). Hyperparameters are selected via grid search over Cartesian products of method-specific parameter grids, and the same search protocol is applied across all experiments, including both main text and Appendix results. For \textbf{DR-SNE}, we sweep $\lambda \in \{0, 5\!\times\!10^{-5}, 10^{-4}, 2\!\times\!10^{-4}, 5\!\times\!10^{-4}, 10^{-3}, 2\!\times\!10^{-3}, 5\!\times\!10^{-3}, 10^{-2}, 10^{-1}, 1.0\}$ and $k_{\text{density}} \in \{40, 80, 300\}$. For \textbf{DenSNE}, we vary perplexity $\in \{5, 10, 15, 30, 50, 75, 100\}$ and $\lambda_{\text{dens}} \in \{0.0, 0.01, 0.05, 0.1, 0.2, 0.5, 1.0\}$, with other parameters fixed (e.g., $\theta=0.5$, $\text{dens\_frac}=0.5$). For \textbf{t-SNE}, we sweep perplexity $\in \{1, 2, 3, 5, 10, 15, 30, 50, 75, 100\}$ and learning rate $\in \{50, 200, 500\}$. For \textbf{UMAP}, we vary $n_{\text{neighbors}} \in \{5, 8, 10, 15, 20, 30, 40, 60, 80, 120\}$ and $\text{min\_dist} \in \{0.0, 0.1, 0.5\}$. For \textbf{DensMAP}, we sweep $\lambda_{\text{dens}} \in \{0.01, 0.05, 0.1, 0.2, 0.5, 1.0, 2.0, 3.0, 5.0, 10.0\}$ and $\text{dens\_frac} \in \{0.1, 0.3, 0.5\}$, with $n_{\text{neighbors}}=30$, $\text{min\_dist}=0.1$, and $\text{densmap}=\text{True}$ fixed. For \textbf{PaCMAP}, we vary $n_{\text{neighbors}} \in \{5, 8, 10, 15, 20, 30, 40, 60, 80, 120\}$ and $\text{FP\_ratio} \in \{1.0, 2.0, 5.0\}$. All results are averaged over 5 random seeds.

\paragraph{Results.}
Table~\ref{tab:anomaly_auprc} summarizes anomaly-detection performance
across embeddings.

On \textbf{Fashion-MNIST}, DR-SNE is competitive but does not achieve the
highest AUPRC for any of the evaluated detectors. Under Oracle-best
selection, DensMAP achieves the highest kNN score, UMAP the highest LOF and
IF scores, and DenSNE the highest centroid score. Under Oracle-IF selection,
DensMAP achieves the highest kNN score, while UMAP performs best for LOF,
IF, and centroid scoring. These results indicate that the downstream utility
of concentration preservation is strongly dataset- and detector-dependent.

On \textbf{PBMC}, DR-SNE performs more strongly. Under Oracle-best
selection, it achieves the highest IF and centroid AUPRC, while UMAP and
DensMAP achieve the highest kNN and LOF scores, respectively. Under
Oracle-IF selection, DR-SNE again achieves the highest IF and centroid
scores, whereas DenSNE performs best for kNN and UMAP for LOF. Thus, no
single embedding method dominates across all detectors, and different
embedding properties appear to benefit different anomaly-scoring rules.

Overall, performance varies substantially across datasets, detectors, and
selection protocols. DR-SNE achieves strong results in several settings,
particularly on PBMC, but does not universally outperform either
density-aware or geometry-focused alternatives.
\clearpage
\section{Effect of Concentration Regularization on Downstream Anomaly Detection}
\label{sec:lambda_anomaly}
\begin{figure*}[!b]
\centering
\includegraphics[width=\textwidth]{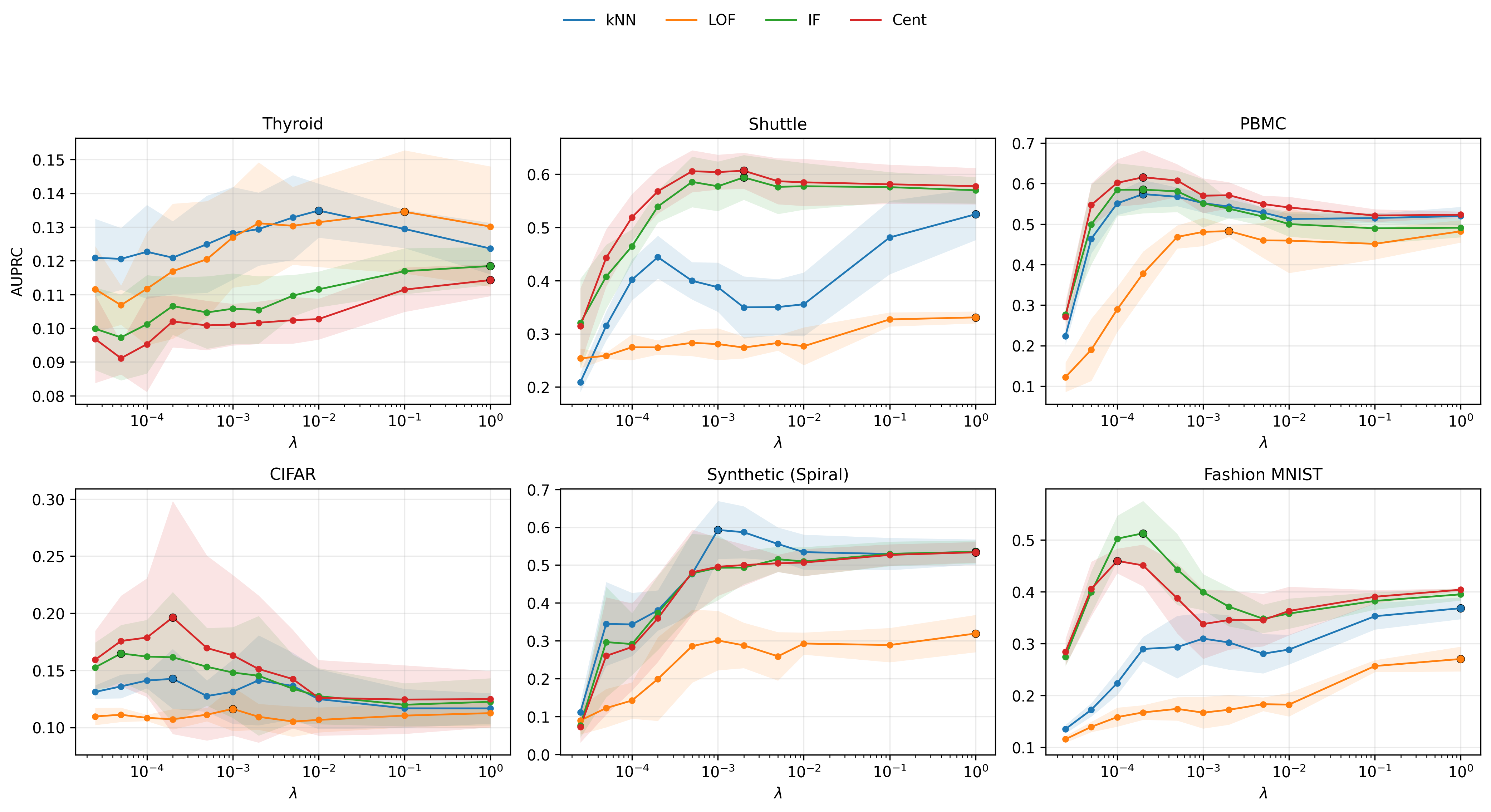}
\caption{
Effect of concentration regularization strength $\lambda$ on anomaly
detection performance (AUPRC). Lines show mean performance across runs;
shaded regions indicate standard deviation.
}
\label{fig:lambda_anomaly}
\end{figure*}

To examine how concentration regularization affects downstream anomaly
detection, we use the same experimental setup, datasets, preprocessing,
anomaly-scoring functions, and AUPRC evaluation protocol as in
Section~\ref{sec:anomaly_detection}. We sweep the concentration
regularization strength $\lambda$ while keeping the remaining DR-SNE
parameters fixed and evaluate the resulting embeddings using kNN, LOF,
Isolation Forest (IF), and centroid-based anomaly scores
(Figure~\ref{fig:lambda_anomaly}).

The effect of concentration regularization is strongly dataset- and
detector-dependent. On several datasets, performance peaks at small but
non-zero values of $\lambda$, typically around $10^{-4}$--$10^{-3}$.
This behavior is particularly evident on PBMC and Fashion-MNIST, where
IF and centroid-based scoring improve substantially under mild
regularization before declining as $\lambda$ increases further. Similar
small-$\lambda$ improvements are observed for several detectors on Shuttle
and CIFAR, although the magnitude and location of the optimum vary across
detectors.

The behavior under stronger regularization is more heterogeneous. On
Thyroid, several detectors improve gradually as $\lambda$ increases, with
LOF in particular attaining its strongest performance at relatively large
$\lambda$. On Shuttle, IF and centroid-based performance increase rapidly
at small $\lambda$ and remain comparatively high across a broad range of
stronger regularization values. Fashion-MNIST also shows partial recovery
for several detectors at large $\lambda$ after an intermediate decline.
Thus, stronger concentration regularization is not uniformly detrimental,
although its effect depends substantially on the dataset and anomaly-scoring
rule.

The correspondence between the regularization sweep and the comparison
against alternative embeddings is particularly informative. On Thyroid,
where stronger concentration regularization remains beneficial across several
detectors, DR-SNE is also among the strongest embedding methods in the
downstream comparison
(Tables~\ref{tab:anomaly_table_three_main}
and~\ref{tab:anomaly_auprc}). By contrast, on datasets where performance is
maximized primarily at small values of $\lambda$, DR-SNE tends to be more
competitive only for particular detectors rather than consistently across
the evaluation.

Overall, these results suggest two broad empirical roles for concentration
regularization. In one regime, the concentration term primarily acts as a
mild regularizer, with small non-zero values of $\lambda$ providing the most
favorable downstream performance. In the other, stronger concentration
preservation remains beneficial, suggesting that the information captured by
the concentration objective may itself be relevant to anomaly detection.

\clearpage
\section{Time Complexity}
\label{sec:time_complexity}

DR-SNE retains the asymptotic complexity of exact t-SNE. Computing pairwise
distances and normalizing the low-dimensional affinities both require
$\mathcal{O}(n^2)$ operations per iteration. The concentration regularizer is
evaluated over $k$-nearest-neighbor neighborhoods and adds only
$\mathcal{O}(nk)$ cost. Thus, the overall per-iteration complexity remains
$\mathcal{O}(n^2)$.

\begin{figure}[!h]
    \centering
    \includegraphics[width=0.45\linewidth]{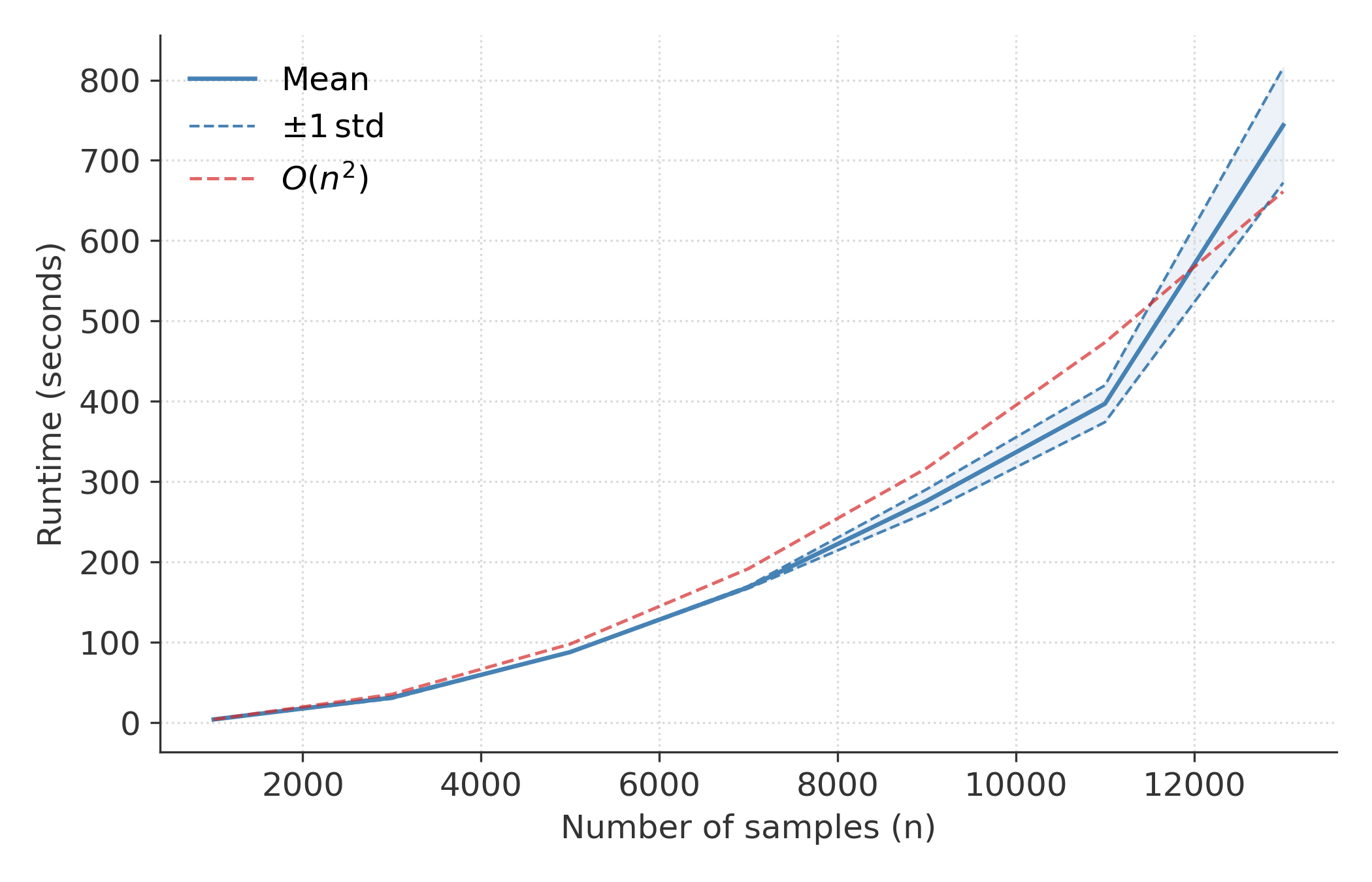}
    \caption{Runtime scaling on Fashion-MNIST. Empirical runtime (blue)
    exhibits approximately quadratic growth, consistent with the
    $\mathcal{O}(n^2)$ reference (red dashed). Shading denotes $\pm 1$
    standard deviation over 7 independent runs. Measurements use
    single-threaded CPU execution.}
    \label{fig:time_complexity}
\end{figure}

Figure~\ref{fig:time_complexity} confirms the expected approximately
quadratic scaling empirically. This limitation is inherited from exact
t-SNE rather than introduced by the concentration regularizer. Standard
t-SNE accelerations reduce the cost to $\mathcal{O}(n\log n)$ with
Barnes--Hut approximations~\citep{van2014accelerating} or to near-linear
scaling with interpolation-based methods such as
FIt-SNE~\citep{linderman2019fast}. Since the DR-SNE regularizer is local and
does not modify the global pairwise interaction structure, it is compatible
in principle with such accelerations; their implementation is left for
future work.

\clearpage
\section{Additional information}

\paragraph{Code availability.}
The implementation of DR-SNE and all experimental pipelines are available at
\url{https://github.com/maksimkazanskii/DR-SNE}.

\paragraph{Use of AI tools.}
AI-assisted tools (ChatGPT, OpenAI GPT-5.3) were used exclusively for language editing and stylistic improvements. They were not used to generate scientific ideas, methods, experiments, or conclusions, all of which were developed independently by the author.

\end{document}